\documentclass[pdflatex,sn-nature]{sn-jnl}

\usepackage{graphicx}%
\usepackage{multirow}%
\usepackage{amsmath,amssymb,amsfonts}%
\usepackage{amsthm}%
\usepackage{mathrsfs}%
\usepackage[title]{appendix}%
\usepackage{xcolor}%
\usepackage{textcomp}%
\usepackage{manyfoot}%
\usepackage{booktabs}%
\usepackage{tabularx}%
\usepackage{colortbl}%
\usepackage[section]{placeins}

\definecolor{oursbg}{RGB}{225,242,225}%
\definecolor{variantbg}{RGB}{229,238,251}%
\newcolumntype{Y}{>{\centering\arraybackslash}X}%
\newcommand{\oursrow}{\rowcolor{oursbg}}%
\newcommand{\varrow}{\rowcolor{variantbg}}%
\newcommand{\grouprow}[2]{\multicolumn{#1}{@{}l}{\textit{#2}}\\}%
\newcommand{\arrdown}{\,$\downarrow$}%
\newcommand{\arrup}{\,$\uparrow$}%
\newcommand{\toone}{\,($\to$1)}%

\newcommand{\method}{CD-FFM}
\newcommand{\degC}{\,$^\circ$C}
\begin{document}

\title[Resolution-free probabilistic climate downscaling]{Resolution-free climate downscaling with likelihood functional flow matching}

\author*[1,2,3]{\fnm{Zhi-Song} \sur{Liu}}\email{zhisong.liu@lut.fi}
\author[2,3]{\fnm{Helmi} \sur{Toropainen}}
\author[1,2,3]{\fnm{Ammar} \sur{Kheder}}
\author[1,2,3]{\fnm{Michael} \sur{Boy}}
\affil[1]{\orgdiv{Department of Computational Engineering}, \orgname{LUT University}, \orgaddress{\city{Lappeenranta}, \country{Finland}}}
\affil[2]{\orgdiv{Institute for Atmospheric and Earth System Research (INAR)}, \orgname{University of Helsinki}, \orgaddress{\city{Helsinki}, \country{Finland}}}
\affil[3]{\orgdiv{Atmospheric Modelling Center, Lahti (AMC-Lahti)}, \orgaddress{\city{Lahti}, \country{Finland}}}

\abstract{Decisions about heat, frost and flooding are taken locally, yet reanalyses and climate models resolve the atmosphere at tens of kilometres. Generative machine learning can add the missing kilometre-scale detail together with its uncertainty, but current generative downscalers are tied to one grid, are costly to sample, produce over-confident ensembles and cannot state how probable an observed field is. We introduce \method{}, a functional flow-matching model that treats the fine-scale field as a random function of location. Because its velocity acts on each location separately, a single trained model can be queried on any grid or at scattered points, the spatial texture of its samples is set by a noise prior fitted to the observed spectrum, and every observation receives an exact likelihood. Downscaling reanalysis to 1\,km daily temperature over Fennoscandia, \method{} is more skilful than regression, a neural operator forty times its size and a diffusion model, while drawing each sample in a fraction of a second.}


\maketitle

\section{Main}\label{sec:main}

Decisions about heat, frost, snow and flooding are taken locally. A road authority decides which stretch of road to salt on a cold night, an energy operator anticipates demand in a city during a heatwave, and an emergency service warns the communities along a single fjord. Climate change is making such local extremes more frequent and harder to anticipate from past experience\cite{fischer2021record}, yet the reanalyses and climate projections that describe the large-scale state of the atmosphere resolve it at roughly 25--100\,km\cite{hersbach2020era5}. In the mountains, valleys and coastlines of northern Europe, the temperature within one such grid cell can differ by several degrees between a valley floor where cold air pools and the plateau above it. Bridging this scale gap, a task known as downscaling, is therefore a prerequisite for translating climate information into local risk assessment and adaptation\cite{maraun2018statistical,wan2024genfocal}.

Existing approaches trade physical fidelity against cost. Dynamical downscaling runs a regional climate model at high resolution; it is physically consistent but too expensive for the large ensembles needed to quantify uncertainty and to sample rare events. Statistical downscaling, from bias correction and spatial disaggregation (BCSD)\cite{wood2004bcsd,maraun2018statistical} to deep regression networks\cite{vandal2017deepsd,rampal2024review}, learns an inexpensive mapping from coarse to fine fields. Most of these methods, however, are deterministic: they return the conditional mean of the fine field and thereby smooth away precisely the fine-scale variability and the tails that matter most for impacts.

Deep generative models have begun to close this gap. Rather than a single mean, they learn the conditional distribution of the fine field and draw ensembles of realistic, kilometre-scale realizations from it. Generative adversarial networks\cite{leinonen2021gan,harris2022gan} were followed by diffusion models\cite{watt2024diffusion,mardani2025corrdiff}, among which CorrDiff\cite{mardani2025corrdiff} corrects a regression network with a residual diffusion model and recovers the power-law spectra of the target data, and GenFocal\cite{wan2024genfocal} combines debiasing and super-resolution to assess regional climate risk from climate projections. Together with the success of generative models in probabilistic weather forecasting\cite{ravuri2021dgmr,price2025gencast}, these works establish that realistic probabilistic downscaling at the kilometre scale is attainable.

Current generative downscalers nevertheless share four limitations. First, they are tied to a grid: the network is trained and sampled on the pixels of one target product, which fixes the resolution and the region at training time. Second, they are costly to sample, typically requiring tens of evaluations of a network with tens of millions of parameters over the whole domain for every ensemble member\cite{karras2022edm,mardani2025corrdiff}, which limits ensemble sizes in exactly the applications that need them. Third, their ensembles are often under-dispersed\cite{mardani2025corrdiff}, so the uncertainty they convey is too narrow. Fourth, their likelihood is intractable at scale: evaluating how probable an observed field is under a diffusion model requires stochastic trace estimates over millions of dimensions\cite{hutchinson1989trace,grathwohl2019ffjord}, so such models are judged by sample-based scores alone. Behind these limitations lies a question that grid-based models leave implicit: what determines the spatial texture of a generated field?

Here we introduce \method{} (Climate Downscaling with Functional Flow Matching), which answers this question by construction. Flow matching\cite{lipman2023flow,liu2023rectified,albergo2023interpolants} trains a network to predict the velocity that transports noise into data, and functional flow matching\cite{kerrigan2024ffm} extends this idea from vectors to random functions. \method{} encodes the coarse reanalysis once and transports a Gaussian random field into the fine-scale field with a velocity that acts on each location separately. This single design choice has three consequences. The model is a random function of location, so that one trained network can be queried at any resolution or at scattered points. The network cannot create spatial structure that is absent from its noise, so the texture of every sample is set by the noise prior, which we fit to the observed spectrum. And at each location the flow is a three-dimensional ordinary differential equation (ODE), so that the likelihood of every observation is exact and cheap. A distilled flow map\cite{boffi2024flowmap} draws a sample in one or two network evaluations, and the whole model has 3.2 million parameters and trains in about an hour on one graphics processor.

We evaluate \method{} by downscaling ERA5\cite{hersbach2020era5} to the 1\,km Nordic Gridded Climate Dataset (NGCD)\cite{ngcd} for daily mean, maximum and minimum temperature over Norway, Sweden and Finland (1.33 million land pixels), with a further experiment on precipitation. Three test years remain unseen during training and model selection. Against the same network trained by regression, a forty-times larger Fourier neural operator\cite{li2021fno} trained as a flow, a BCSD-style statistical baseline and a CorrDiff-style diffusion model, \method{} gives the most skilful and best-calibrated probabilistic downscaling in every test year, reproduces the observed spatial texture in flat and in mountainous terrain, and samples about fifty times faster than the diffusion model.

\section{\method}\label{sec:model}

\begin{figure}[!htbp]
\centering
\includegraphics[width=\textwidth]{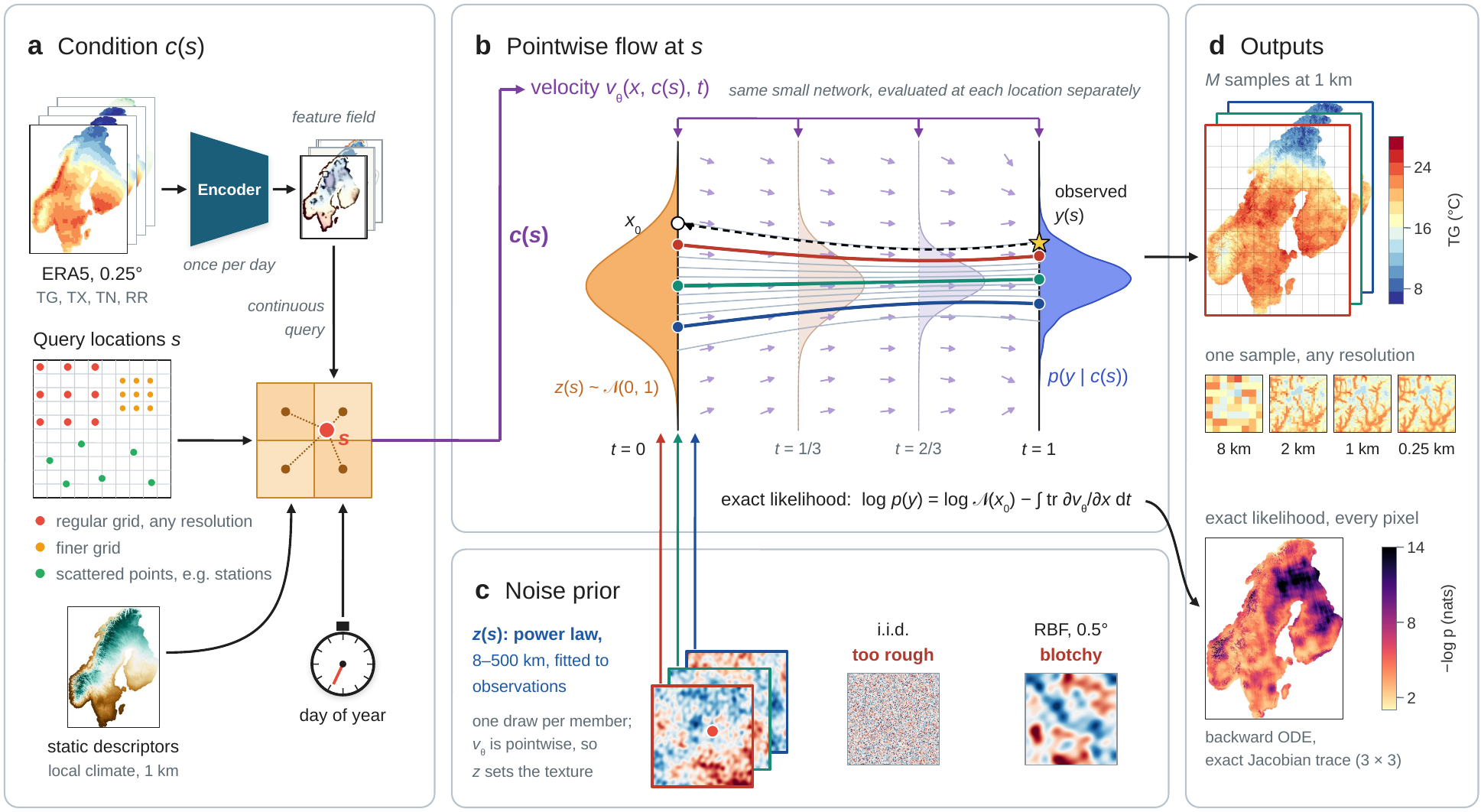}
\caption{\textbf{\method{} downscales by querying a random function of location.} \textbf{a}, The ERA5 fields of one day (TG, TX, TN and RR at 0.25$^\circ$) are encoded once into a coarse feature field. Any set of query locations $s$ --- a regular grid at any resolution, a finer grid or scattered points such as stations --- is read from it by a continuous query that combines the four surrounding cells with the sub-cell offset, the static descriptors of the local climate at 1\,km and the day of year, giving a condition $c(s)$. \textbf{b}, At each location, a small network $v_\theta(x, c(s), t)$ transports the noise value $z(s)\sim\mathcal{N}(0,1)$ at flow time $t=0$ into the predictive distribution $p(y\,|\,c(s))$ at $t=1$. Streamlines and light arrows show the exact flow-matching velocity for a one-dimensional predictive distribution shaped like that of the daily minimum temperature at a pixel near Troms{\o} on 15 January 2019, with the intermediate densities at $t=1/3$ and $2/3$ in light colours; bold paths are three ensemble members. Integrating the same ordinary differential equation backwards from an observation $y(s)$ (star) to $x_0$ gives its exact likelihood. \textbf{c}, Each member $m$ starts from its own draw $z^{(m)}(s)$ (coloured lines to $t=0$ in \textbf{b}) of a Gaussian random field whose power spectrum is fitted to the observations (8--500\,km). Because the velocity cannot move information between locations, this field sets the texture of the sample; i.i.d.\ noise would be too rough and a smooth radial-basis-function (RBF) field blotchy. \textbf{d}, Outputs for 28 July 2019: three 1\,km samples of daily mean temperature (frames coloured as the members in \textbf{b} and \textbf{c}), one sample queried at 8, 2, 1 and 0.25\,km, and the exact negative log-likelihood of the observed (TG, TX, TN) at every pixel, obtained with the exact trace of a 3$\times$3 Jacobian rather than a stochastic estimator. Thumbnails in \textbf{a} show the ERA5 input and static descriptors of the same day.}
\label{fig:schematic}
\end{figure}

Figure~\ref{fig:schematic} summarizes the model (Methods). For a given day, an encoder\cite{flower} maps the ERA5 fields at 0.25$^\circ$ --- daily mean, maximum and minimum 2\,m temperature and total precipitation --- to a coarse feature field. At any query location $s$, a local implicit function\cite{chen2021liif} reads the features of the four surrounding ERA5 cells together with the sub-cell offset and 17 static descriptors of the local climate, and returns a condition $c(s)$. A small network then defines the velocity $v_\theta(x, c(s), t)$ that transports the noise value $z(s)$ at $t=0$ into the daily mean, maximum and minimum temperature (TG, TX and TN) at $t=1$. The targets are standardized anomalies with respect to the local NGCD climatology, so that the network learns how a day departs from what is expected at a place and season rather than the seasonal cycle itself.

Because $v_\theta$ depends on location only through $c(s)$, the joint distribution of the outputs at any finite set of points is the push-forward of the noise field through independent per-point maps. The distributions of different point sets are therefore consistent with one another, and \method{} defines a single stochastic process rather than a separate prediction for each grid (Supplementary Information~\ref{si:consistency}). The same argument shows that training on random subsets of points is an unbiased estimate of the process-level objective, which is what makes training on scattered points fast. It also exposes a less obvious property. A pointwise map can transform the noise value at each location, but it cannot move information between locations; given the condition, the spatial dependence of a sample is inherited from $z$. Choosing the kernel of $z$ thus amounts to choosing the texture of the output, and we treat it as a modelling decision in its own right: instead of the usual white or smooth noise, we draw $z$ from a Gaussian random field whose power spectrum follows the slope measured on the NGCD residual field (Methods).

The pointwise design also makes the likelihood tractable. At each pixel, the predictive density of the observed (TG, TX, TN) is a three-dimensional Gaussian pushed through the learned ODE. It follows exactly from the instantaneous change-of-variables formula\cite{chen2018node,grathwohl2019ffjord} at the cost of one backward ODE solve, with the divergence computed exactly as the trace of a 3$\times$3 Jacobian rather than estimated stochastically. Finally, to sample in one or two steps rather than ten, we distil the learned flow into a flow map\cite{boffi2024flowmap,frans2025shortcut} that jumps between any two times of the ODE in one network evaluation (Methods).

\section{One reanalysis day, many plausible kilometre-scale fields at any resolution}\label{sec:overview}

\begin{figure}[!htbp]
\centering
\includegraphics[width=\textwidth]{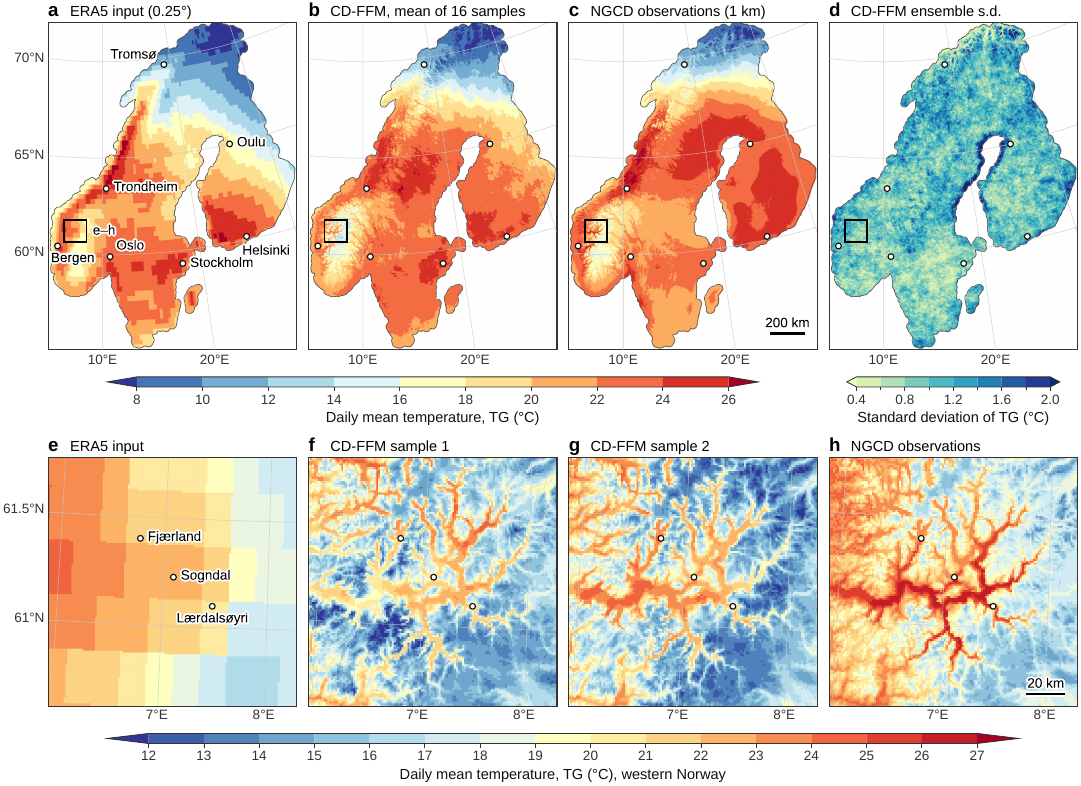}
\caption{\textbf{\method{} turns one coarse ERA5 day into an ensemble of plausible 1\,km temperature fields.} \textbf{a}--\textbf{d}, Daily mean temperature (TG) on 28 July 2019, a summer day of the test year, over Norway, Sweden and Finland: ERA5 input at 0.25$^\circ$ (\textbf{a}), the mean of 16 \method{} samples (\textbf{b}), NGCD observations at 1\,km (\textbf{c}) and the ensemble standard deviation (\textbf{d}). The square marks the 128$\times$128\,km region in western Norway enlarged in \textbf{e}--\textbf{h}. \textbf{e}--\textbf{h}, ERA5 input (\textbf{e}), two individual \method{} samples (\textbf{f}, \textbf{g}) and NGCD observations (\textbf{h}) in that region, a landscape of fjords, valleys and high plateaus around Sogndal. Samples are drawn with the distilled flow map in two steps.}
\label{fig:overview}
\end{figure}

Figure~\ref{fig:overview} follows a summer day of the test year 2019, 28 July, when a heatwave reached northern Fennoscandia. The ERA5 input resolves the broad north--south contrast but none of the fjords and valleys of western Norway. Each \method{} sample adds this structure: within the 0.25$^\circ$ cells around Sogndal, warm valley floors alternate with cold plateaus as in the NGCD observations, while the samples differ from one another in the local detail (Fig.~\ref{fig:overview}e--h and Extended Data Fig.~\ref{fig:ed_ensemble}). Averaged over the domain, a single sample departs from NGCD by 1.83\degC{} (mean absolute error) and the mean of 16 samples by 1.55\degC, against 3.02\degC{} for the ERA5 field itself. In the ensemble mean the sample-specific details cancel (Fig.~\ref{fig:overview}b), whereas the ensemble standard deviation, 1.0\degC{} on average, shows where the reanalysis leaves the local value uncertain: it is largest in the north and inland and smallest along the well-predicted southern coasts (Fig.~\ref{fig:overview}d).

Because \method{} is a function of continuous location, the same sample can be evaluated on any set of points. Querying a single noise realization on nested grids from 16 to 1\,km over the whole domain, or from 8 to 0.25\,km over a mountain box, yields values that agree at shared points to within $2\times10^{-5}$\degC{} (Extended Data Fig.~\ref{fig:ed_multires}): coarse and fine queries are views of one field rather than separate predictions. The cost scales with the number of points queried. A full 1\,km field of the domain takes 0.94\,s on one GPU for the first member, including encoding, and about 0.35\,s for each further member; a 4\,km field takes 0.06\,s and a 16\,km field 0.01\,s. Below the 1\,km training resolution, down to 0.25\,km, the field remains continuous and free of blocks or seams. This finer detail is a smooth continuation of what the encoder, the static descriptors and the noise resolve, not new information, because no observations finer than 1\,km constrain it.

\section{The noise prior sets the texture of generated fields}\label{sec:noise}

\begin{figure}[!htbp]
\centering
\includegraphics[width=\textwidth]{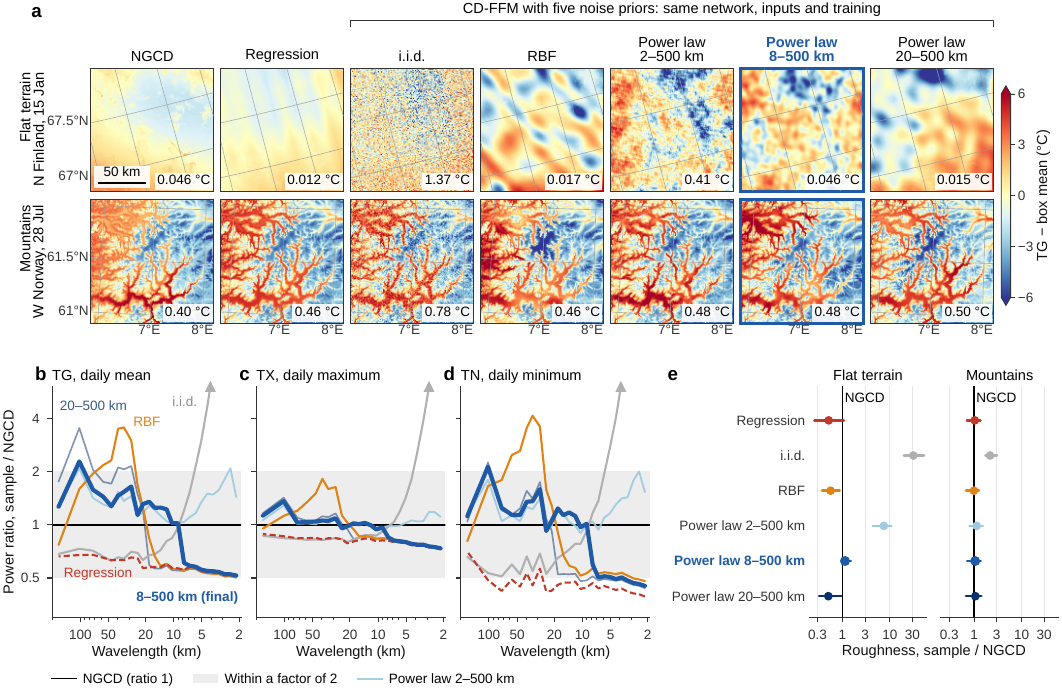}
\caption{\textbf{A noise prior fitted to the observed spectrum reproduces the texture of the observations; standard priors do not.} \textbf{a}, One sample of daily mean temperature minus its box mean in a 128$\times$128\,km box of flat terrain in northern Finland on 15 January 2019 (top) and of mountainous terrain in western Norway on 28 July 2019 (bottom): NGCD, the regression network and \method{} trained with five noise priors that differ in nothing else (i.i.d., RBF, and power laws over 2--500, 8--500 and 20--500\,km; blue frame, final configuration). The number in each panel is the pixel-scale roughness (mean absolute deviation from the 3$\times$3 median). \textbf{b}--\textbf{d}, Power spectrum of one sample divided by that of NGCD for TG (\textbf{b}), TX (\textbf{c}) and TN (\textbf{d}), averaged over every seventh day of 2019 (53 days) in a 512$\times$512\,km box in southern Norway and Sweden; the shaded band marks agreement within a factor of two, and arrows mark the i.i.d.\ prior leaving the axis (13--47 times the observed power at 2\,km). \textbf{e}, Pixel-scale roughness relative to NGCD in flat and mountainous terrain: dot, mean over 15 January, 28 July and 6 November 2019; bar, range over the three days. NGCD roughness on these days is 0.046, 0.031 and 0.030\degC{} in flat terrain and 0.59, 0.40 and 0.32\degC{} in the mountains.}
\label{fig:texture}
\end{figure}

That a pointwise flow inherits its spatial texture from the noise is not only a theoretical statement. To isolate the effect, we trained five models that differ only in the noise prior and compared their samples with NGCD in flat and in mountainous terrain (Fig.~\ref{fig:texture}). Spatially independent (i.i.d.) noise produces salt-and-pepper fields: in flat terrain their pixel-scale roughness is about 30 times that observed, and at a wavelength of 2\,km they carry 13--47 times the observed power. A conventional smooth random field has the opposite defect: with a squared-exponential kernel of 0.5$^\circ$ length scale, samples carry two to four times too much power at 30--150\,km, visible as blotches, and about half the observed power below 15\,km. In both cases the signature is that of the kernel, which shows directly that the network passes the noise texture through to the output.

We therefore draw the noise from a power-law spectrum $S(k)\propto k^{-\beta}$ with $\beta=2.4$, the slope of the NGCD residual spectrum at 5--100\,km (Methods). The range of wavelengths that this power law covers turns out to matter. Spanning 2--500\,km, it makes flat terrain far too rough: in northern Finland, the pixel-scale roughness of a sample is 0.41 and 0.14\degC{} on a winter and a summer day, against 0.046 and 0.031\degC{} in NGCD, because the observed fine-scale variability is concentrated in the mountains, whereas a stationary prior spreads it uniformly. Restricting the prior to 8--500\,km removes this artefact. On the same days the samples have the roughness of NGCD in flat terrain (0.046 and 0.030\degC), and at every wavelength resolved in a 512\,km box (2--170\,km) their spectrum stays within a factor of about two of the observed one: the power ratio ranges from 0.50 to 2.3 for TG, 0.71 to 1.36 for TX and 0.44 to 2.1 for TN (Fig.~\ref{fig:texture}b--d). Below 8\,km the samples carry about half the observed power of TG and TN, and restricting the prior further, to 20--500\,km, makes them too smooth in flat terrain (roughness about half that observed). In mountainous terrain, where the static descriptors supply most of the fine-scale structure, every prior except i.i.d.\ gives a roughness within 0.7--1.4 times that of NGCD (Fig.~\ref{fig:texture}e), so flat terrain is the discriminating case. Even there the final prior is not perfect: on a third test day (6 November) its samples are 1.5 times too rough, and on the winter day their variability at 5--20\,km is visibly larger than observed (Fig.~\ref{fig:texture}a), which may partly reflect the smoothness of NGCD where stations are sparse.

Per-pixel scores are nearly blind to these differences. The continuous ranked probability scores (CRPS) of the five priors differ by at most 0.08\degC{} for TG and 0.008\degC{} for TX and TN, and the smoothest prior (20--500\,km) even scores best for TG (Extended Data Table~\ref{tab:ed_noise}). Such scores evaluate each pixel separately and cannot see that a sample is too smooth or too rough. Applications that use the generated fields as fields --- areal means, catchment totals, spatial extremes, inputs to hydrological or impact models --- need the spatial structure to be right, which is why we select the prior on spectra and roughness and use the 8--500\,km power law, the only prior whose flat-terrain roughness is close to that observed on average (1.16 times, range 0.97--1.51 over three days), in all further experiments.

\section{Skilful probabilistic downscaling in every test year}\label{sec:skill}

\begin{figure}[!htbp]
\centering
\includegraphics[width=\textwidth]{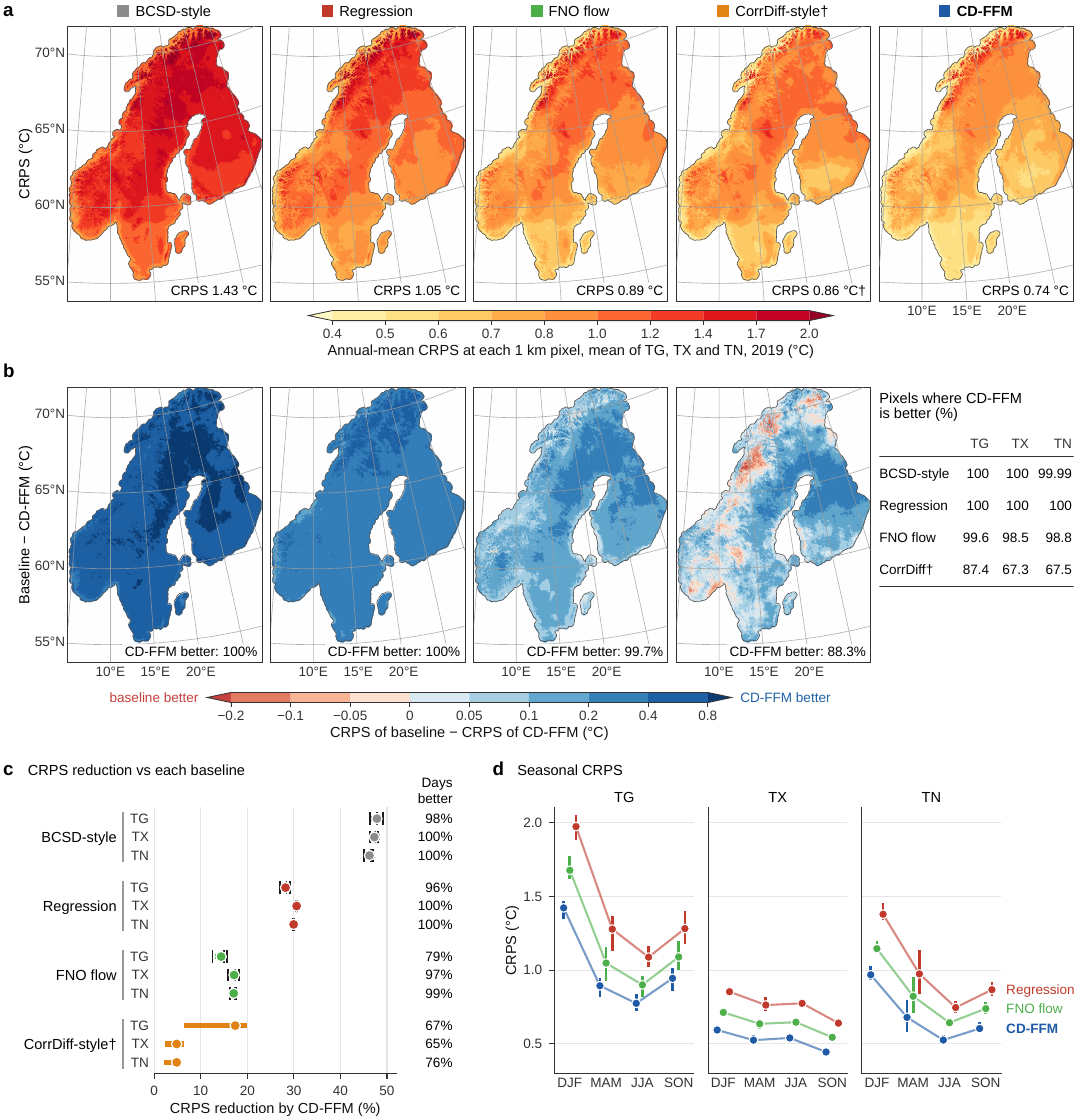}
\caption{\textbf{\method{} has the lowest probabilistic error almost everywhere and in every test year.} \textbf{a}, Annual-mean fair CRPS at each 1\,km pixel (mean of TG, TX and TN), test year 2019; the label gives the domain mean. \textbf{b}, CRPS of each baseline minus that of \method{} on the same days (blue: \method{} better), with the share of pixels at which \method{} is better, on each map and per variable in the table. \textbf{c}, CRPS reduction of \method{} relative to each baseline, 2016, 2018 and 2019 pooled (1{,}096 days): dot, estimate; bar, 95\% paired moving-block bootstrap interval (7-day blocks; may be hidden by the dot); ticks, single test years; right, percentage of days on which \method{} is better. \textbf{d}, Seasonal CRPS pooled over the three test years (dots), with the range of the single-year values (bars). \method{}: flow map, two steps, 16 members. $^\dagger$CorrDiff-style model: every seventh day of 2019 (55 days), 8 members, paired with \method{} on the same days and with the same number of members (\method{} domain mean on those days: 0.76\degC).}
\label{fig:skill}
\end{figure}

We next compare probabilistic skill with that of four baselines trained on the same 15 years, inputs and evaluation code (Table~\ref{tab:main}, Fig.~\ref{fig:skill} and Methods). None of the three test years was used for training or checkpoint selection. \method{} has a lower CRPS than the regression, neural-operator and BCSD-style baselines for every variable and in every test year. Pooled over 2016, 2018 and 2019 (1{,}096 days), the reduction is 28--31\% relative to the same network trained by regression (27--31\% in single years), on 96--100\% of the days, and 14--17\% relative to a Fourier neural operator (FNO) with 124 million parameters trained as a flow with identical inputs and noise (12.5--18\% in single years), on 79--99\% of the days; all 95\% block-bootstrap intervals\cite{kunsch1989bootstrap} exclude zero (Fig.~\ref{fig:skill}c and Extended Data Table~\ref{tab:ci}). The gain is spatially uniform: averaged over the test year and over the three variables, \method{} has a lower CRPS than regression and the BCSD-style baseline at every one of the 1.33 million pixels, and than the FNO flow at 99.7\% of them (Fig.~\ref{fig:skill}a,b). It also holds in every season: \method{} has the lowest seasonal CRPS in all 36 combinations of season, test year and variable, with the largest absolute gain in winter, when all models err most (Fig.~\ref{fig:skill}d).

\begin{table}[!htbp]
\caption{\textbf{\method{} is the most skilful model on every probabilistic score.} Test year 2019, all 1.33 million pixels, best validation checkpoint (chosen on 2017). CRPS: fair ensemble CRPS (equal to the mean absolute error for deterministic models). RMSE of the ensemble mean. Spread/skill: ensemble spread over RMSE of the ensemble mean (1 indicates calibration); Brier scores of extremes are given in Extended Data Table~\ref{tab:ed_extremes}. $^\dagger$Every seventh day of 2019 (55 days) with 8 members, because of the sampling cost of the diffusion model; these two rows are compared only with each other. Bold: best value among the models evaluated on every day, and the better model in the $\dagger$ block. Green: final \method{} configuration; blue: other \method{} samplers.}
\label{tab:main}
\footnotesize
\setlength{\tabcolsep}{3pt}%
\begin{tabularx}{\textwidth}{@{}lcYYYYYYYYY@{}}
\toprule
\multirow{2}{*}{\textbf{Model}} & \multirow{2}{*}{\textbf{Params}} & \multicolumn{3}{c}{\textbf{CRPS} (\degC)\arrdown} & \multicolumn{3}{c}{\textbf{RMSE} (\degC)\arrdown} & \multicolumn{3}{c}{\textbf{Spread/skill}\toone} \\
\cmidrule(lr){3-5}\cmidrule(lr){6-8}\cmidrule(l){9-11}
 & & TG & TX & TN & TG & TX & TN & TG & TX & TN \\
\midrule
\grouprow{11}{Generative models, every day, 16 members}
FNO flow & 124M & 1.195 & 0.639 & 0.837 & \textbf{1.90} & 1.08 & 1.49 & 0.17 & 0.28 & 0.27 \\
\varrow \method, 10 Euler steps & 3.2M & 1.034 & 0.530 & 0.699 & 1.97 & 1.05 & 1.48 & 0.51 & 0.61 & 0.59 \\
\oursrow \textbf{\method, flow map} & 3.2M & \textbf{1.008} & \textbf{0.522} & \textbf{0.689} & 1.97 & 1.06 & 1.49 & \textbf{0.61} & \textbf{0.75} & \textbf{0.72} \\
\midrule
\grouprow{11}{Deterministic models, every day}
BCSD-style & --- & 1.984 & 1.005 & 1.302 & 2.71 & 1.41 & 1.83 & --- & --- & --- \\
CorrDiff-style UNet & 24.8M & 1.616 & 0.787 & 1.026 & 2.22 & 1.11 & 1.51 & --- & --- & --- \\
Regression & 3.2M & 1.422 & 0.752 & 0.986 & 1.98 & 1.06 & 1.47 & --- & --- & --- \\
\varrow \method, zero noise & 3.2M & 1.386 & 0.727 & 0.959 & 1.94 & \textbf{1.04} & \textbf{1.44} & --- & --- & --- \\
\midrule
\grouprow{11}{Every seventh day, 8 members$^\dagger$}
CorrDiff-style & 24.8M & 1.288 & 0.549 & 0.730 & 2.36 & 1.10 & \textbf{1.48} & 0.40 & 0.64 & 0.62 \\
\oursrow \textbf{\method, flow map} & 3.2M & \textbf{1.064} & \textbf{0.523} & \textbf{0.694} & \textbf{2.08} & \textbf{1.08} & 1.50 & \textbf{0.55} & \textbf{0.70} & \textbf{0.69} \\
\botrule
\end{tabularx}
\end{table}

The gain is probabilistic rather than deterministic. The error of the \method{} ensemble mean is within $\pm$1.2\% of that of the regression network, as expected for a model trained to match distributions rather than to minimize the error of its mean, and the near-deterministic FNO ensemble mean is slightly more accurate for TG (mean absolute error 1.34 against 1.41\degC). What the FNO lacks is spread: its members are nearly identical (spread/skill ratio 0.17--0.28, where 1 indicates calibration), so it loses on every probabilistic score. \method{} is the best calibrated of all generative models, with spread/skill ratios of 0.61, 0.75 and 0.72 for TG, TX and TN, although it remains under-dispersed (Table~\ref{tab:main} and Extended Data Fig.~\ref{fig:ed_rank}).

To test \method{} against the strongest class of generative downscalers, we re-implemented the two-stage CorrDiff design\cite{mardani2025corrdiff}: a UNet regression of the mean followed by an EDM\cite{karras2022edm} residual diffusion model, run on the 1\,km grid with exactly the inputs of \method{} (Methods). To avoid disadvantaging this baseline, we gave it about ten times the training compute of \method{} (13.1 against 1.4 GPU-hours, the latter including distillation). Because full-domain diffusion sampling is expensive, both models were compared with 8 members on every seventh day of 2019. \method{} is more skilful for every variable, reducing CRPS by 17, 5 and 5\% for TG, TX and TN, with paired bootstrap intervals that exclude zero, and it is better at 88\% of the pixels (87\% for TG and 67\% for TX and TN; Fig.~\ref{fig:skill}b). The diffusion model is competitive for TX and TN, where its calibration approaches that of \method{}, and clearly worse for TG, where its ensemble is more under-dispersed (spread/skill 0.40 against 0.55) and its ensemble mean less accurate (RMSE 2.36 against 2.08\degC). \method{} achieves this at a fraction of the cost: each diffusion sample requires 35 evaluations of a 24.8-million-parameter UNet, which in the same evaluation pipeline took about 19\,s per member and day, against 0.35\,s for the \method{} flow map.

\section{Local extremes and their probabilities}\label{sec:extremes}

\begin{figure}[!htbp]
\centering
\includegraphics[width=\textwidth]{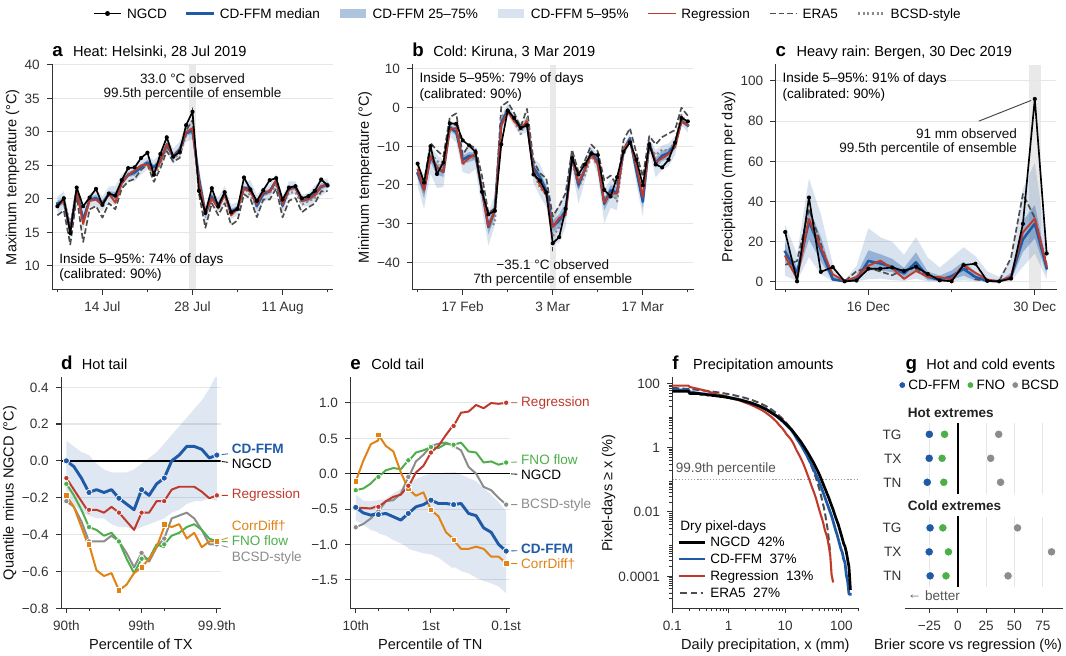}
\caption{\textbf{\method{} reproduces the hot and heavy-precipitation tails most faithfully, but its ensembles are too narrow on the most extreme days.} \textbf{a}--\textbf{c}, Daily values at the 1\,km pixel of one of eleven Nordic cities around the most extreme day of 2019, selected from NGCD alone: the highest daily maximum temperature (Helsinki, \textbf{a}), the lowest daily minimum temperature (Kiruna, \textbf{b}) and the largest daily precipitation (Bergen, \textbf{c}); windows cover $\pm$21 days, and \textbf{c} ends on 31 December. Black, NGCD; blue, median and 25--75\% and 5--95\% ranges of 200 \method{} samples; red, regression; dashed, ERA5 (enclosing 0.25$^\circ$ cell for temperature, bilinear for precipitation); dotted, BCSD-style baseline. The grey column marks the event day; labels give the percentile of the observation among the samples on that day, and boxes the fraction of days on which the observation lies inside the 5--95\% range (90\% if calibrated). \textbf{d}, \textbf{e}, Quantiles of one sample per pixel-day minus those of NGCD, for the upper tail of the daily maximum (\textbf{d}) and the lower tail of the daily minimum temperature (\textbf{e}), pooled over 20{,}000 fixed land pixels and every day of 2019; shading shows 95\% moving-block bootstrap intervals over days. \textbf{f}, Fraction of the same pixel-days with daily precipitation of at least $x$; the legend gives the dry fraction ($<0.1$\,mm) and the dotted line marks the 99.9th percentile. \textbf{g}, Brier score of the ensemble probability of hot (above the local 95th percentile) and cold (below the 5th) days relative to the regression network, all pixels and days of 2019. $^\dagger$CorrDiff-style model: every seventh day of 2019 only (55 days).}
\label{fig:extremes}
\end{figure}

For decisions, the most consequential output of a downscaler is the probability that a local threshold will be crossed. We score it with the Brier score of the ensemble probability that a value exceeds the local 95th percentile of the seasonal NGCD climatology, or falls below its 5th percentile. Relative to regression, \method{} reduces the Brier score by 24--27\% for hot and cold extremes of all three variables, about twice the reduction achieved by the FNO flow (Fig.~\ref{fig:extremes}g), and on the days on which both were evaluated its Brier scores are lower than those of the CorrDiff-style model for all hot extremes and cold TG, and equal for cold TX and TN (Extended Data Table~\ref{tab:ed_extremes}).

Three extremes selected from the observations alone show both what the ensemble adds and where it falls short (Fig.~\ref{fig:extremes}a--c). On 28 July 2019, Helsinki reached a daily maximum of 33.0\degC; the \method{} median (30.2\degC) was as low as the regression prediction (30.5\degC), and only 0.5\% of the samples reached the observed value. At Kiruna on 3 March, the observed minimum of $-35.1$\degC{} lay at the 7th percentile of the ensemble. In Bergen, the 91\,mm observed on 30 December exceeded all but 0.5\% of the samples (median 29\,mm), whereas the regression network and ERA5 gave 31 and 32\,mm and no indication that such an amount was possible. Over the six weeks around the two temperature events and the three weeks around the precipitation event, the observation lay within the central 90\% of the ensemble on 74, 79 and 91\% of the days, against 90\% for a calibrated ensemble: the temperature ensembles are too narrow during these extremes, consistent with their overall under-dispersion.

Pooled over 20{,}000 fixed pixels and every day of 2019, \method{} reproduces the upper tail of the daily maximum temperature best: its 99th and 99.9th percentiles differ from those of NGCD by $-0.16$ and $+0.03$\degC, against $-0.2$ to $-0.6$\degC{} for the other methods (Fig.~\ref{fig:extremes}d). In the lower tail of the daily minimum, the picture is mixed: at the 0.1st percentile \method{} is too cold by 1.1\degC{} and the regression network too warm by 1.0\degC, and on the 55 days on which both were evaluated the cold tail of the CorrDiff-style model is closer to NGCD than that of \method{} (1.3 against 2.8\degC{} too cold; Fig.~\ref{fig:extremes}e). For precipitation, \method{} comes closest to the observed fraction of dry pixel-days (37\% against 42\%; regression 13\%, ERA5 27\%) and to the heavy tail (99.9th percentile 38.9 against 44.5\,mm; regression 27.1\,mm, ERA5 36.1\,mm), and it is the only method whose largest value, 151\,mm, reaches the observed maximum of 150\,mm (Fig.~\ref{fig:extremes}f).

\section{An exact likelihood for every observation}\label{sec:likelihood}

Sample-based scores say how close an ensemble is to the observations; a likelihood says how probable the observations are under the model, and it can be evaluated, mapped and compared. Because the flow at each location is a three-dimensional ODE, \method{} provides this quantity exactly, at every pixel and without sampling (Methods). Over every seventh day of 2019, it assigns the observed TG, TX and TN the highest likelihood of all models we can score: 4.71 nats per pixel in negative log-likelihood, against 5.10 for the regression network and 7.78--9.12 for the FNO flows, whose near-deterministic ensembles place almost no probability on the observed values. The regression score is optimistic, because we give that network the most favourable Gaussian spread, fitted on the test days themselves. Even so, the observed values are on average about 1.5 times more probable under \method{}.

\begin{figure}[!htbp]
\centering
\includegraphics[width=\textwidth]{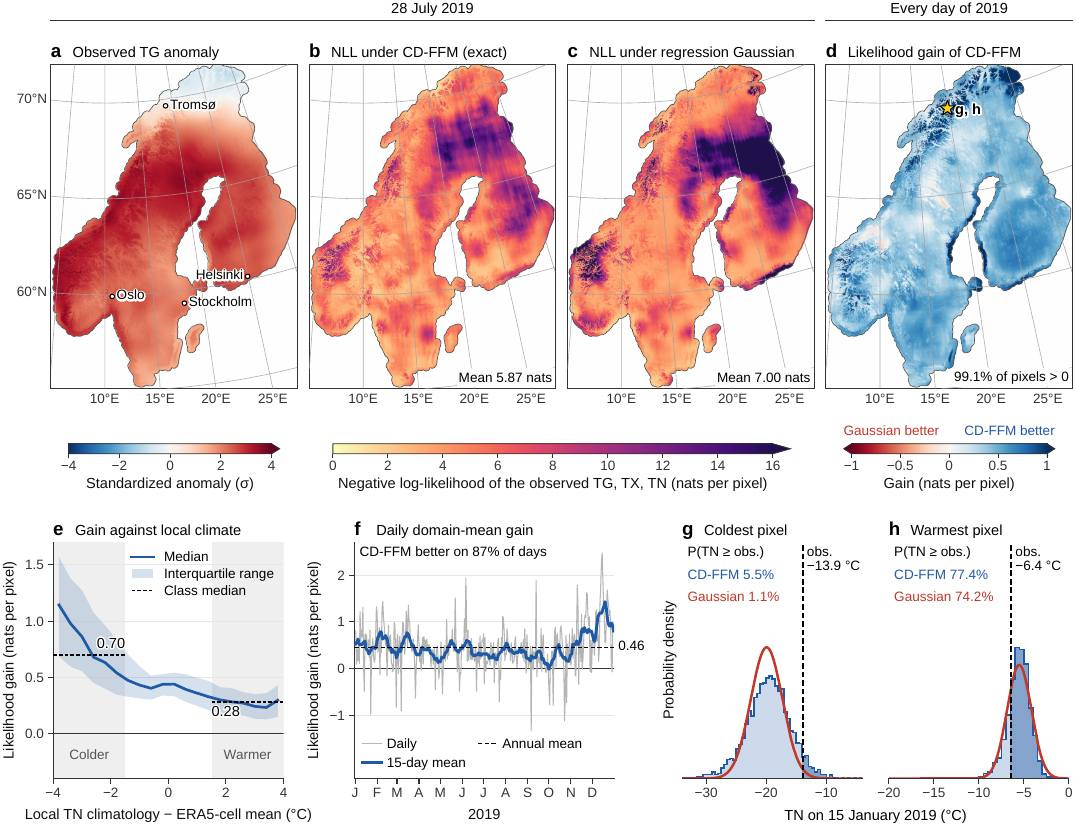}
\caption{\textbf{The exact likelihood shows how surprising each observation is, and where non-Gaussian local distributions matter.} \textbf{a}, Observed daily mean temperature on 28 July 2019 as a standardized anomaly. \textbf{b}, \textbf{c}, Negative log-likelihood (NLL) of the observed (TG, TX, TN) at every pixel on that day under \method{} (exact) and under the regression network with its most favourable Gaussian (residual covariance fitted on the test days); boxes give the domain means. \textbf{d}, Likelihood gain of \method{} over the Gaussian (Gaussian NLL minus \method{} NLL), averaged over every day of 2019; positive values mean that the observations are more probable under \method{}. The star marks the pixel of \textbf{g}. \textbf{e}, Gain against the deviation of the local climatological minimum temperature from the mean of its ERA5 cell (negative: sites colder than their cell, such as valley floors): median and interquartile range in 0.4\degC{} bins, with the medians of the two shaded classes. \textbf{f}, Daily domain-mean gain over 2019 (grey) and its 15-day running mean (blue); dashed, annual mean. \textbf{g}, \textbf{h}, Predictive distribution of the minimum temperature on 15 January 2019 at the climatologically coldest (\textbf{g}; $-6.6$\degC{} relative to the cell mean) and warmest (\textbf{h}; $+5.0$\degC) pixel of the ERA5 cell with the largest local contrast, near Troms{\o}: 4{,}000 \method{} samples (histogram), the regression Gaussian (red) and the observation (dashed), with the probability of values at least as warm as observed.}
\label{fig:likelihood}
\end{figure}

Evaluated on every day of 2019, the observations are more probable under \method{} than under the regression Gaussian at 99.1\% of the pixels, with an annual mean gain of 0.46 nats per pixel, and on 87\% of the days (Fig.~\ref{fig:likelihood}d,f). The gain depends on the local climate. It is largest at sites whose climatological minimum temperature lies more than 1.5\degC{} below the mean of their ERA5 cell, typically valley floors where cold air pools (median 0.70 nats), and smallest at sites more than 1.5\degC{} above it (0.28 nats; Fig.~\ref{fig:likelihood}e). At such sites the minimum temperature given the coarse field is far less certain --- an inversion may form or break --- and a Gaussian with a single standardized covariance for all pixels cannot represent this local width. At the coldest pixel of the cell with the largest local contrast, near Troms{\o}, the observed minimum on 15 January 2019 was $-13.9$\degC, about 6\degC{} above both predictive medians; \method{} gave 5.5\% probability to values at least this warm, the Gaussian 1.1\% (Fig.~\ref{fig:likelihood}g).

Mapped over the domain, the likelihood shows where the observed field is surprising given the reanalysis (Fig.~\ref{fig:likelihood}a--c). On 28 July 2019, the least likely observations lie in the interior of northern Sweden and Finland, where the heat was least expected given the reanalysis (mean negative log-likelihood that day 5.87 nats per pixel, against 4.63 over the year), whereas the well-predicted south has values of about 2--4 nats per pixel. Such maps, which take about 15\,s per day for the 1.33 million pixels on one GPU, can flag unusual local weather or problems in the observational product at every location. Differentiating the log-likelihood of one observation with respect to the coarse input further shows where the model draws its information from (Extended Data Fig.~\ref{fig:ed_influence}). Half of the gradient mass of \method{} lies within 360--490\,km of the target, in an anisotropic, location-specific pattern, whereas the FNO flow spreads its sensitivity almost uniformly over the domain (half the mass within 690--850\,km). The same figure shows that one-point correlation maps of \method{} samples reproduce the observed spatial dependence of daily anomalies, whereas the regression output correlates too strongly over long distances, as expected from a model that predicts smooth conditional means.

\section{From research model to practical tool}\label{sec:practice}

Several properties determine whether a downscaler can be used routinely. We summarize them here and give details in Extended Data and Supplementary Information.

\textbf{One-step sampling.} Integrating the learned ODE with a coarse Euler scheme trades accuracy for speed: with two steps the ensemble contracts (spread/skill 0.17) and CRPS rises by 16--20\% relative to ten steps. The distilled flow map removes this trade-off (Extended Data Fig.~\ref{fig:steps}). One flow-map step is as accurate as twenty Euler steps (CRPS 1.067, 0.524 and 0.696\degC, against 1.077, 0.524 and 0.697\degC) and better calibrated, and two steps are 1.3--2.5\% more skilful than the ten-step sampler for every variable, on 88--95\% of the days of the three test years (Extended Data Table~\ref{tab:ci}). The likely reason is that the map was trained on accurate integrations of the teacher ODE, whereas a coarse integrator under-integrates the flow and contracts the ensemble.

\textbf{A best estimate from the same model.} Many users need a single map. Integrating the learned ODE from zero noise gives a smooth deterministic field (Extended Data Fig.~\ref{fig:ed_smooth}) that is more accurate than the regression network trained with the same inputs (mean absolute error 1.39, 0.73 and 0.96\degC, against 1.42, 0.75 and 0.99\degC) and 7--14\% more accurate than the 24.8-million-parameter UNet regression stage of the CorrDiff-style baseline. One trained model thus provides both the best estimate and the samples around it.

\textbf{Consistency with a biased coarse field.} One might expect a downscaled field to average back to its coarse input, but ERA5 is not a coarsening of NGCD: their daily 28\,km cell means differ by about 2\degC{} on average. Forcing the samples to reproduce the ERA5 cell means exactly raises the CRPS of TG by 60\%, whereas the unconstrained model produces cell means that are closer to NGCD than ERA5 itself is (mean absolute difference 1.61 against 1.98\degC; Extended Data Table~\ref{tab:ed_infer}). \method{} thus performs bias correction and downscaling jointly, and hard conservation constraints, which suit a coarse field that aggregates the target, are harmful when the coarse field is biased.

\textbf{Locations without training data.} Because \method{} is a function of location, generalization can be tested by never sampling a region during training. Withholding northern Fennoscandia (about 132{,}000 pixels) changes the CRPS in that region by $-0.03$, $+0.03$ and $+0.06$\degC{} for TG, TX and TN: no detectable effect for TG and a small penalty of 4 and 7\% for TX and TN (Extended Data Table~\ref{tab:ed_holdout} and Supplementary Information~\ref{si:further}). The model still uses the climatology of the region for normalization, so this experiment tests generalization to unseen daily data at locations with a known climatology.

\textbf{Precipitation.} Trained unchanged except for the transform of amounts to a Gaussian scale, \method{} downscales daily precipitation with a CRPS of 1.17\,mm\,day$^{-1}$, against 1.65 for regression and 2.21 for interpolated ERA5, reproduces the frequency of wet days (frequency bias 1.09, against 1.50 for regression) and of days above 10\,mm (0.94, against 0.52), and has the lowest Brier score at every threshold (Extended Data Table~\ref{tab:precip}). It produces too few events above 20\,mm (frequency bias 0.79), and its samples are more fragmented than observed, because the noise prior was fitted to temperature; since the velocity is pointwise, a coarser noise band can be chosen at sampling time without changing any per-pixel score (Extended Data Fig.~\ref{fig:ed_precip_tricks}).

\textbf{Cost.} The model reaches its best validation score after about one hour of training on a single AMD Instinct MI250X graphics compute die, and distillation adds 27 minutes. More training data helps up to about ten years (Extended Data Fig.~\ref{fig:ed_scaling}), and the static descriptors, which tell the model whether a pixel lies in a valley, on a slope or at the coast, lower the CRPS of TX and TN by 3 and 5\% (Extended Data Table~\ref{tab:ed_ablation}).

\section{Discussion}\label{sec:discussion}

\method{} recasts generative downscaling as learning a random function of location. A pointwise velocity on top of a continuous implicit representation removes the grid from the model, keeps the network that runs at each location small, and makes three things possible that grid-based generative downscalers do not offer together: one model that can be queried at any resolution or set of points, samples whose texture is controlled by a measurable noise prior, and the exact likelihood of every observation. Across three test years, this design is more skilful than regression, statistical downscaling and a much larger neural-operator flow on every probabilistic score, and more skilful than a CorrDiff-style diffusion model while sampling about fifty times faster.

The results suggest that, for downscaling, the design of the noise deserves as much attention as the design of the network. Diffusion and flow models usually start from i.i.d.\ Gaussian noise; for a model that acts on each location separately this choice fixes a white-noise texture, and a conventional smooth kernel imposes its own characteristic scale. Measuring the residual spectrum of the target and drawing the noise from it is a simple, physically interpretable alternative that per-pixel scores alone would never select. That no single stationary prior fits both flat and steep terrain points to the next step, a non-stationary prior whose spectrum varies with terrain or weather regime. Because the flow is invertible, such a prior could be estimated rather than designed: integrating observed fields backwards yields the noise that would have produced them.

The exact likelihood turns a generative downscaler into a probabilistic model whose density can be evaluated, mapped and compared. In this study it ranked models and noise priors, revealed where the local predictive distribution departs from a Gaussian, and located the observations that the reanalysis does not explain. The same quantity could serve as a model-selection criterion, as a routine quality check for gridded observational products, and as a basis for combining downscaled fields with station observations at arbitrary locations, which the resolution-free design makes possible without regridding.

Our study has limitations. The ensembles remain under-dispersed (spread/skill 0.61--0.75), as for other generative downscalers\cite{mardani2025corrdiff}, most visibly during extremes, and the extreme cold tail is not yet reliable: at 20{,}000 fixed pixels over 2019, 413 of 7.3 million sampled pixel-days fall below $-40$\degC{} (down to $-59.5$\degC), values that never occur in NGCD, so a bounded or physically constrained output transform is needed before the coldest samples are used. The stationary noise prior underestimates fine-scale variability in steep terrain and, averaged over the domain, adds too much variability within the 0.25$^\circ$ cells: the interquartile range of the departures of TG from its cell mean is 2.4 times that observed (Supplementary Information~\ref{si:further}). The precipitation model reuses the prior fitted to temperature and under-produces the heaviest events. \method{} relies on the NGCD climatology for normalization and static descriptors and is therefore not yet applicable where no high-resolution observational record exists, although a global kilometre-scale climatology and elevation data could replace both. We evaluate one region and one target product, which is itself an interpolation of station data, and our CorrDiff-style baseline is a re-implementation trained with our inputs and evaluated on a subset of days for cost reasons.

The efficiency of \method{} opens the way to large ensembles of kilometre-scale fields, such as downscaled ensemble forecasts and long climate projections in which rare local extremes can only be sampled with many members\cite{wan2024genfocal}. Whether the learned mapping remains valid in the warmer climates of such projections is an open question that requires dedicated tests. More broadly, we hope that treating the noise prior as a measurable and designable part of a generative model will prove useful well beyond downscaling.

\section{Methods}\label{sec:methods}

\subsection*{Data}

\textbf{Target.} NGCD\cite{ngcd} provides daily TG, TX, TN and precipitation (RR) on a 1\,km Lambert azimuthal equal-area grid (2{,}020$\times$1{,}550 cells) over Norway, Sweden and Finland, interpolated from station observations\cite{lussana2019senorge}. We use the 1{,}325{,}781 land pixels with complete climatologies for all three temperature variables.

\textbf{Input.} ERA5\cite{hersbach2020era5} daily mean, maximum and minimum 2\,m temperature and total precipitation of the target day on a 0.25$^\circ$ grid (71$\times$113 cells, 54--71.5$^\circ$N, 4--32$^\circ$E). ERA5 temperatures are standardized with a two-harmonic seasonal climatology fitted per cell; precipitation is log-transformed and standardized globally.

\textbf{Splits.} We train on 2001--2015 (5{,}478 days), select checkpoints on 2017 (12 fixed dates, 20{,}000 fixed pixels) and test on 2016, 2018 and 2019. Normalization statistics are fitted on 1981--2015.

\textbf{Normalization.} Each temperature variable is standardized per pixel as $z = (x - \mu(s,d))/\sigma(s,d)$, where $\mu$ and $\sigma^2$ are two-harmonic seasonal fits of the mean and of the residual variance, $d$ is the day of year, and $\sigma$ is floored at 0.5\degC. Precipitation is mapped to a Gaussian scale through its cumulative probability $F$: a dry day ($\leq0.1$\,mm) maps to $F=p_\mathrm{dry}(s)$, the per-pixel dry-day frequency, and a wet amount to $F = p_\mathrm{dry} + (1-p_\mathrm{dry})\,G(x)$, where $G$ is a global quantile table (5--95\%) above which a generalized Pareto distribution (threshold 15.6\,mm) describes the tail; in both cases $z=\Phi^{-1}(F)$.

\textbf{Static descriptors.} Seventeen per-pixel features computed from the 1981--2015 NGCD climatology act as a terrain proxy: for each temperature variable, the annual mean, its deviation from the mean of the enclosing ERA5 cell, the amplitude of the annual cycle and the January and July standard deviations, together with the TX$-$TN diurnal range and its sub-cell deviation. The precipitation model adds $p_\mathrm{dry}$ and its sub-cell deviation.

\subsection*{Model}

\textbf{Encoder and continuous query.} A U-shaped Flower encoder\cite{flower} with warping-based mixing (four resolution levels, lifting dimension 64, 8 heads, 8 groups) maps the four standardized ERA5 fields of the day to a 128-channel feature field at ERA5 resolution. For a location $s$ we follow the local implicit image function\cite{chen2021liif} with feature unfolding (3$\times$3 neighbourhoods) and local ensembling over the four surrounding cells, weighted by the opposite-corner area, with the static descriptors concatenated at each of the four positions. A three-layer perceptron returns the 64-dimensional condition $c(s)$, to which a day-of-year encoding is appended.

\textbf{Velocity.} $v_\theta(x, c(s), t)$ is a four-layer perceptron (width 256, GELU) on the concatenation of the state $x\in\mathbb{R}^3$, $c(s)$ and a sinusoidal embedding of $t$, applied to every location independently. The full model has 3.2 million parameters.

\textbf{Noise prior.} We construct $z$ from random Fourier features\cite{rahimi2007rff}, $z(s) = \sqrt{2/F}\sum_{f=1}^{F} a_f \cos(2\pi\,\omega_f\!\cdot\!\tilde{s} + \varphi_f)$, with $a_f\sim\mathcal{N}(0,1)$ and $\varphi_f\sim\mathcal{U}(0,2\pi)$ drawn independently for each output channel, where $\tilde{s}$ is the location in degrees with longitude scaled by $\cos 64.5^\circ$; $z$ has unit marginal variance. For the power-law prior, the frequencies $\omega_f$ have uniform direction and radial density $\propto S(k)\,k = k^{1-\beta}$ on $[k_0, k_1]$, drawn by inverting the cumulative distribution, so that the expected spectrum is $S(k)\propto k^{-\beta}$ between wavelengths $1/k_1$ and $1/k_0$. We use $\beta = 2.4$, the slope of the NGCD residual spectrum (observations minus regression) at 5--100\,km for all three variables, wavelengths of 8--500\,km and $F=1{,}024$. On a 512$\times$512\,km test grid this construction recovers spectral slopes of $-2.39$ for $\beta = 2.4$ and $-3.21$ for $\beta = 3.0$. The RBF prior uses a squared-exponential kernel with a 0.5$^\circ$ length scale and 64 features.

\textbf{Training.} We use flow matching\cite{lipman2023flow,liu2023rectified,albergo2023interpolants} with the linear interpolant $x_t=(1-t)z+tx_1$ and the loss $\mathbb{E}\,\|v_\theta(x_t,c,t)-(x_1-z)\|^2$, masked where NGCD is missing. Each batch contains 16 days with 2{,}560 points per day: 2{,}048 points drawn uniformly over the training region and 16 clusters of 32 points inside randomly chosen ERA5 cells, so that sub-cell contrasts are seen in every batch. We optimize with AdamW\cite{loshchilov2019adamw} (learning rate $3\times10^{-4}$, 1{,}000 warm-up steps, cosine decay, gradient clipping at 1) and keep an exponential moving average of the weights (decay 0.999). Training runs for 60{,}000 steps with checkpoints every 5{,}000 steps; the evaluated checkpoint minimizes the summed validation CRPS and is reached after about 40{,}000 steps. All training data reside on the GPU as 16-bit standardized arrays, so a training step takes about 85\,ms.

\textbf{Flow map.} A second network $u_\eta(x, c, t, s)$ has the architecture of $v_\theta$ plus a sinusoidal embedding of $s-t$ and defines the map $\Psi(x_t,t,s) = x_t + (s-t)\,u_\eta(x_t,c,t,s)$. It is initialized from $v_\theta$ with the new inputs zeroed, so that it starts as one Euler step. With the encoder and query frozen, it is trained to regress the teacher's average velocity $(x_s - x_t)/(s-t)$, where $x_s$ is obtained from $x_t$ by integrating $v_\theta$ with eight fourth-order Runge--Kutta substeps; on 25\% of the points we set $s=t$ and use $v_\theta(x_t,t)$ as the target, so that the instantaneous velocity is retained. Distillation runs for 20{,}000 steps with a learning rate of $10^{-4}$ (27\,min). Unless stated otherwise, results use the flow map with two steps and 16 members.

\textbf{Arbitrary-resolution queries.} Query points are placed on corner-aligned nested grids in the index space of the NGCD grid, so that every point of a coarser grid also belongs to all finer ones. Coordinates, static descriptors and climatology coefficients are interpolated bilinearly in grid-index space; the ERA5 encoding is computed once, and the noise field, the condition and the flow are evaluated at the points with a fixed random seed.

\textbf{Exact likelihood.} Because $v_\theta$ is pointwise and $z(s)\sim\mathcal{N}(0,I_3)$ at every pixel, the predictive density of $(\mathrm{TG},\mathrm{TX},\mathrm{TN})$ at a pixel is $\mathcal{N}(0,I_3)$ pushed through a three-dimensional ODE. By the instantaneous change of variables\cite{chen2018node,grathwohl2019ffjord}, $\log p_1(x_1) = \log\mathcal{N}(x_0;0,I_3) - \int_0^1 \nabla_x\!\cdot v_\theta(x_t,c,t)\,\mathrm{d}t$, where $x_t$ is obtained by integrating the ODE backwards from the observation. The divergence is the trace of a 3$\times$3 Jacobian, computed exactly with three backward passes, so no stochastic trace estimator\cite{hutchinson1989trace} is needed and the difficulty addressed by recent likelihood flow maps\cite{ouyang2026scallop} does not arise. We use 20 fourth-order Runge--Kutta steps (10, 20 and 40 steps agree to 0.001 nats) and convert to physical units by adding $-\sum_c\log\sigma_c(s,d)$. Deterministic models receive a Gaussian with the 3$\times$3 residual covariance fitted on the evaluated days themselves, an oracle that gives them the most favourable homoscedastic spread. For influence maps we evaluate the same quantity differentiably and back-propagate it through the velocity, the query and the encoder to the standardized ERA5 input.

\subsection*{Baselines}

\textbf{Regression.} The same network and inputs, with $v$ evaluated at $x=0$, $t=0$ as a direct predictor trained with mean squared error in standardized space. \textbf{FNO flow.} The Flower encoder replaced by an FNO encoder\cite{li2021fno} in the same U-shaped scaffold, at lifting dimensions 64 (124M parameters) and 16 (7.7M), with everything else identical. \textbf{BCSD-style.} The ERA5 standardized anomaly interpolated bilinearly and rescaled with the NGCD climatology, $\hat{x} = \mu(s,d) + \sigma(s,d)\,z_\mathrm{ERA5}(s)$. \textbf{CorrDiff-style.} Following ref.~\cite{mardani2025corrdiff}, a UNet regression of the standardized mean followed by an EDM\cite{karras2022edm} residual diffusion model on the 1\,km grid, with Karras preconditioning, log-normal noise levels and a deterministic Heun sampler with 18 steps (35 network evaluations). Both UNets (channel multipliers 1-2-4-4, base 64, two residual blocks per level, no attention; 24.8M parameters) receive the inputs of \method{} as 1\,km channels and are trained on random 128$\times$128\,km patches; inference is tile-wise with overlapping tiles, as in the original (Supplementary Information~\ref{si:baselines}). \textbf{Precipitation baseline.} ERA5 total precipitation interpolated bilinearly in log space.

\subsection*{Evaluation}

All scores are computed on every land pixel of every test day. \textbf{CRPS} uses the fair ensemble estimator\cite{ferro2014fair}, $\frac{1}{M}\sum_i|x_i-y| - \frac{1}{2M(M-1)}\sum_{i,j}|x_i-x_j|$, which is unbiased for the CRPS of the underlying distribution and therefore comparable across ensemble sizes and with deterministic predictions\cite{gneiting2007scoring}. \textbf{Spread/skill} is the mean ensemble standard deviation divided by the RMSE of the ensemble mean, and \textbf{rank histograms}\cite{hamill2001rank} use 16 members. \textbf{Extremes} are scored with the Brier score of the ensemble probability of exceeding the local 95th percentile ($z>1.645$) or falling below the 5th ($z<-1.645$) of the seasonal climatology; for precipitation we use absolute thresholds of 0.1, 1, 10 and 20\,mm\,day$^{-1}$ with frequency bias and Brier score. \textbf{Spectra} are radially averaged power spectra of one sample in the largest fully valid 512$\times$512\,km box (southern Norway and Sweden), every seventh day. \textbf{Roughness} is the mean absolute deviation from the 3$\times$3 median in 128$\times$128\,km boxes in northern Finland (flat) and western Norway (mountains). \textbf{Significance.} Daily domain means of each score are paired between models on common days, and the mean difference is resampled with a moving-block bootstrap\cite{kunsch1989bootstrap} (7-day blocks within each year, 4{,}000 resamples); we report the 2.5th and 97.5th percentiles.

\textbf{Compute.} All experiments ran on single AMD Instinct MI250X graphics compute dies (64\,GB) of the LUMI supercomputer, in PyTorch. All experiments in this paper, including discarded configurations and the CorrDiff-style baseline, used about 110 GPU-hours.

\backmatter

\bmhead{Data availability}
ERA5 is available from the Copernicus Climate Change Service Climate Data Store. NGCD is available from the Copernicus Climate Change Service (\url{https://doi.org/10.24381/cds.e8f4a10c}). The preprocessed inputs and targets of the test years, the climatologies and static descriptors, the weights of all models in this paper, and the evaluation outputs behind every table and figure are available at Zenodo (\url{https://doi.org/10.5281/zenodo.23496724}).

\bmhead{Code availability}
Code for preprocessing, training, evaluation and all figures, the training configurations of every model and pretrained weights are available at \url{https://github.com/Holmes-Alan/CD-FFM} and archived at Zenodo (\url{https://doi.org/10.5281/zenodo.23499076}) under the MIT licence.

\bmhead{Acknowledgements}
We acknowledge the EuroHPC Joint Undertaking for awarding this project access to the LUMI supercomputer.

\bmhead{Author contributions}
Z.-S.L. conceived the method, implemented the models and experiments, and drafted the manuscript. H.T. collected data and revised the manuscript. M.B. advised on the atmospheric interpretation and revised the manuscript.

\bmhead{Competing interests}
The authors declare no competing interests.

\begin{appendices}

\renewcommand{\figurename}{Extended Data Fig.}%
\renewcommand{\tablename}{Extended Data Table}%
\renewcommand{\thefigure}{\arabic{figure}}%
\renewcommand{\thetable}{\arabic{table}}%
\renewcommand{\theHfigure}{ED\arabic{figure}}%
\renewcommand{\theHtable}{ED\arabic{table}}%
\setcounter{figure}{0}%
\setcounter{table}{0}%

\section*{Extended Data}\label{sec:ed}

\begin{figure}[!htbp]
\centering
\includegraphics[width=\textwidth]{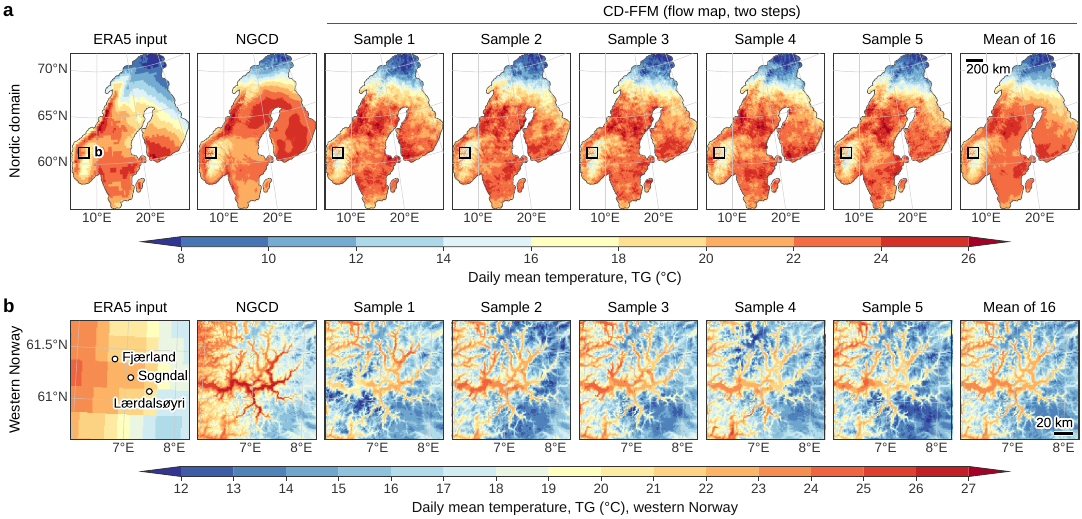}
\caption{\textbf{One input, many plausible 1\,km fields.} \textbf{a}, Daily mean temperature on 28 July 2019 over the Nordic domain: ERA5 input, NGCD observations, five \method{} samples (flow map, two steps) and the mean of 16 samples. \textbf{b}, The same in the 128$\times$128\,km region in western Norway. The samples share the large-scale pattern set by ERA5 and differ in the local detail.}
\label{fig:ed_ensemble}
\end{figure}

\begin{figure}[!htbp]
\centering
\includegraphics[width=\textwidth]{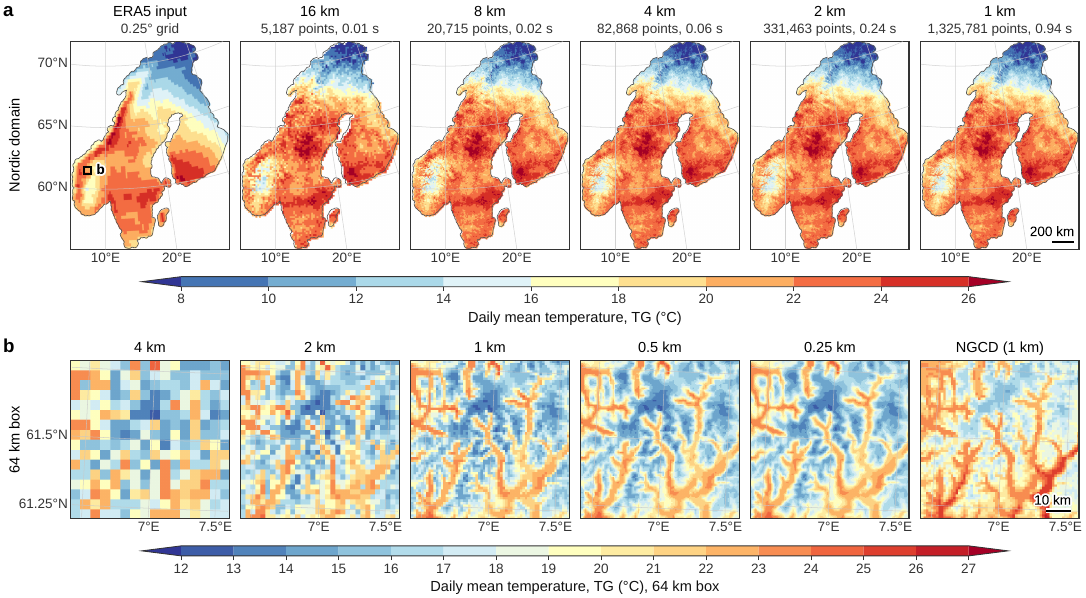}
\caption{\textbf{One sample at any resolution over the whole domain.} \textbf{a}, ERA5 input and one \method{} sample (fixed noise realization) queried separately at 16, 8, 4, 2 and 1\,km over the Nordic domain, with the number of points and the GPU time of each query (encoding, query and two flow-map steps). \textbf{b}, A 64$\times$64\,km box in western Norway (outlined in \textbf{a}) queried at 4, 2, 1, 0.5 and 0.25\,km, and NGCD at 1\,km. Values at shared points agree to $2\times10^{-5}$\degC. Coarse grids are point values, not area averages, which is why they alternate between valley and mountain in steep terrain.}
\label{fig:ed_multires}
\end{figure}

\begin{figure}[!htbp]
\centering
\includegraphics[width=\textwidth]{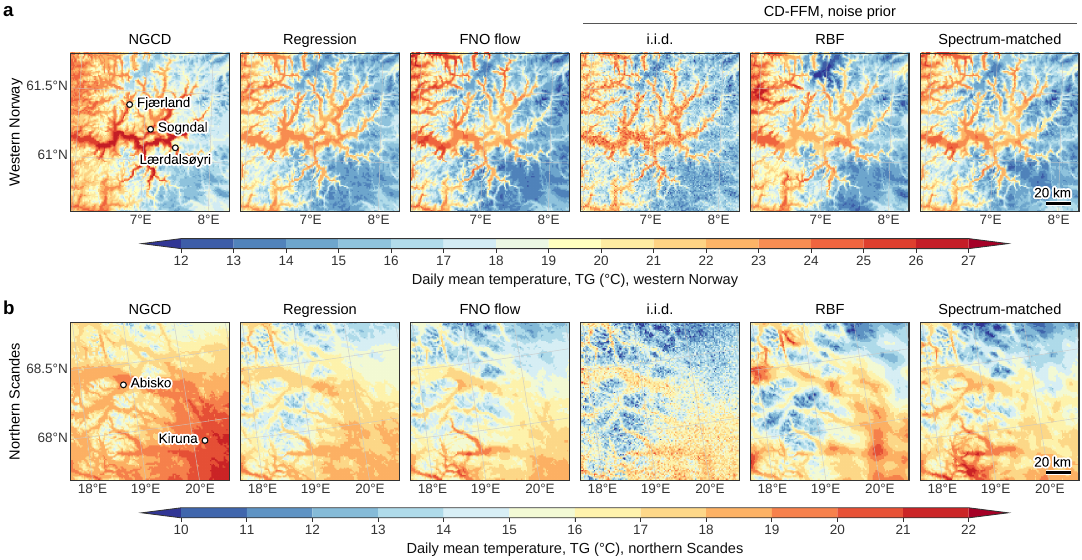}
\caption{\textbf{Generated samples reproduce the fjord and mountain structure of NGCD.} Daily mean temperature on 28 July 2019 in two 128$\times$128\,km regions, western Norway (\textbf{a}) and the northern Scandes (\textbf{b}): NGCD observations, the regression network, and one sample each of the FNO flow and of \method{} with i.i.d., RBF and spectrum-matched (8--500\,km) noise (all \method{} samples from the learned ODE with ten Euler steps, so that only the noise prior differs). Each generative panel is a single sample, whose departure from the observations includes the ensemble spread and any error of the ERA5 input that all methods share; the panels compare spatial structure, whereas accuracy is compared in Fig.~\ref{fig:skill} and Table~\ref{tab:main}.}
\label{fig:ed_maps}
\end{figure}

\begin{figure}[!htbp]
\centering
\includegraphics[width=\textwidth]{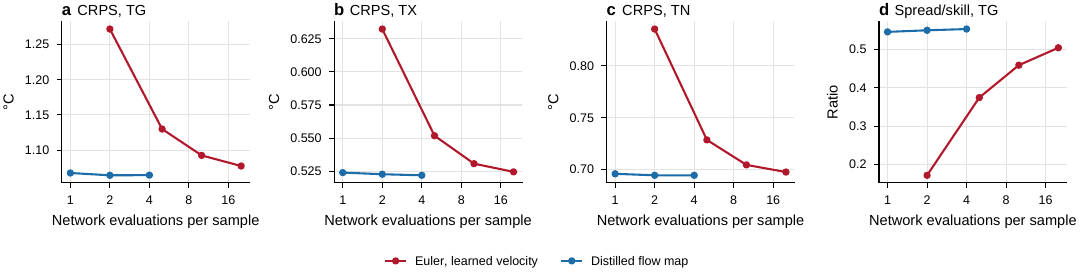}
\caption{\textbf{A distilled flow map is as skilful in one network evaluation as the ODE in twenty.} \textbf{a}--\textbf{c}, CRPS of TG, TX and TN against the number of network evaluations per sample, for Euler integration of the learned velocity (red) and for the distilled flow map (blue). \textbf{d}, Spread/skill ratio for TG (1 indicates calibration). Every seventh day of 2019, 8 members.}
\label{fig:steps}
\end{figure}

\begin{figure}[!htbp]
\centering
\includegraphics[width=\textwidth]{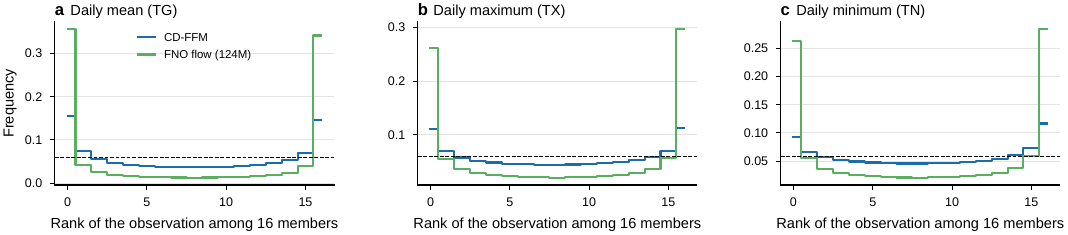}
\caption{\textbf{\method{} is better calibrated than the FNO flow but still under-dispersed.} \textbf{a}--\textbf{c}, Rank histograms of the observed TG, TX and TN among 16 members, 2019; the dashed line is the calibrated level.}
\label{fig:ed_rank}
\end{figure}

\begin{figure}[!htbp]
\centering
\includegraphics[width=\textwidth]{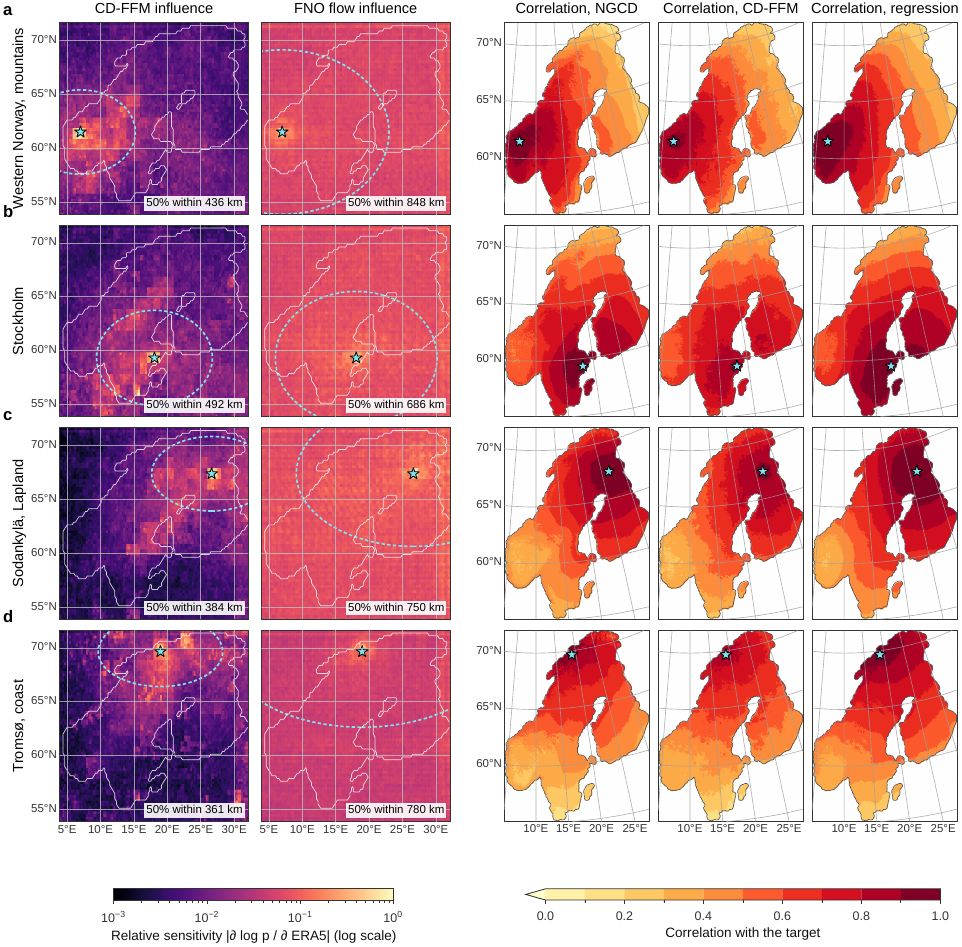}
\caption{\textbf{\method{} draws information from a compact, location-specific region and reproduces the observed spatial dependence.} \textbf{a}--\textbf{d}, Four target locations (star). Columns 1--2, likelihood influence maps $|\partial \log p(y(s^*)\,|\,\mathrm{ERA5}) / \partial\,\mathrm{ERA5}|$, summed over the four input fields and averaged over every 14th day of 2019, for \method{} and the FNO flow, normalized by their maximum (log colour scale); the dashed circle and the label give the radius holding 50\% of the gradient mass; white line, land. These gradients are local sensitivities, not causal effects. Columns 3--5, one-point correlation maps of the daily TG anomaly at the target with every 1\,km pixel over 2019: NGCD observations, one \method{} sample per day, regression.}
\label{fig:ed_influence}
\end{figure}

\begin{figure}[!htbp]
\centering
\includegraphics[width=\textwidth]{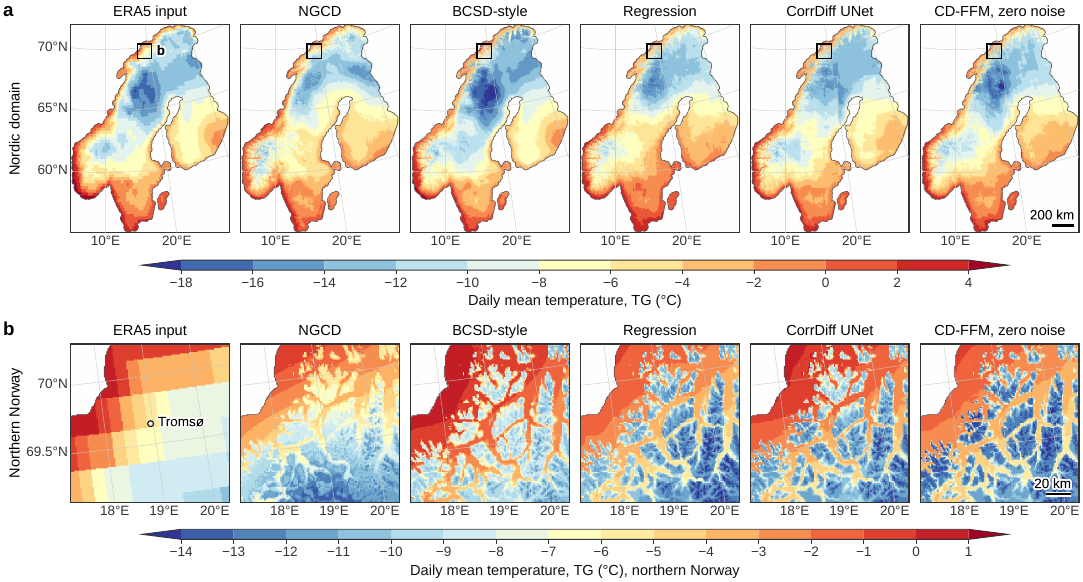}
\caption{\textbf{The zero-noise ODE gives a smooth best estimate.} \textbf{a}, Daily mean temperature on 15 January 2019 over the Nordic domain: ERA5 input, NGCD observations, the BCSD-style baseline, the regression network with the same architecture, the UNet regression stage of the CorrDiff-style baseline (tile-wise inference) and the zero-noise \method{} estimate. \textbf{b}, The same in a 128$\times$128\,km region in northern Norway around Troms{\o}. The zero-noise estimate is as smooth as the regression outputs and, over the test year, more accurate than both (Table~\ref{tab:main}); on this single day the regression network is closer to NGCD.}
\label{fig:ed_smooth}
\end{figure}

\begin{figure}[!htbp]
\centering
\includegraphics[width=0.5\textwidth]{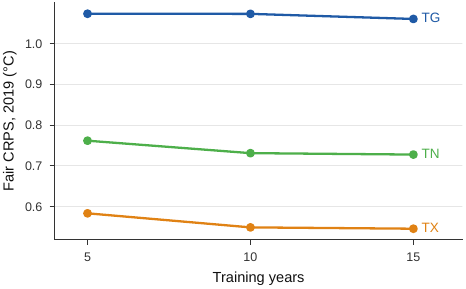}
\caption{\textbf{More training data helps up to about ten years.} Fair CRPS on 2019 for models trained on the last 5, 10 and 15 years of the training period (2011--2015, 2006--2015 and 2001--2015), with RBF noise and without static descriptors, otherwise identical. Going from 5 to 10 years lowers CRPS by 6 and 4\% for TX and TN; going from 10 to 15 years lowers it by less than 1\%.}
\label{fig:ed_scaling}
\end{figure}

\begin{figure}[!htbp]
\centering
\includegraphics[width=\textwidth]{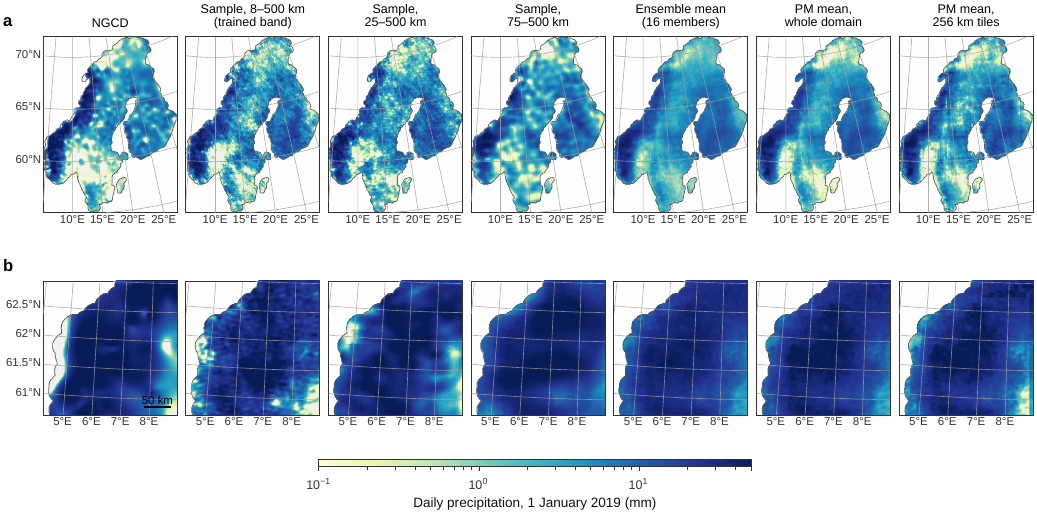}
\caption{\textbf{Precipitation texture can be corrected at sampling time.} Daily precipitation on 1 January 2019 over the Nordic domain (\textbf{a}) and in a 256$\times$256\,km box in western Norway (\textbf{b}), log colour scale (grey: dry): NGCD; one sample with the trained noise band (8--500\,km) and with coarser bands (25--500 and 75--500\,km) used only at sampling time; the 16-member ensemble mean; and probability-matched means\cite{ebert2001pm} over the whole domain and over 256\,km tiles. Per-pixel CRPS and Brier scores are the same for all three bands (2019, every seventh day). The tile-wise matching leaves seams and is not recommended.}
\label{fig:ed_precip_tricks}
\end{figure}

\begin{table}[!htbp]
\caption{\textbf{The gains are significant and consistent across years.} Paired differences in daily domain-mean CRPS (\degC; A $-$ B, negative favours A), pooled over 2016, 2018 and 2019 (1{,}096 days), with 95\% moving-block bootstrap intervals (7-day blocks within years) and the fraction of days on which A scores better. A is the \method{} flow map (2 steps, 16 members). Every interval excludes zero. Blue: comparison with another \method{} sampler. $^\dagger$2019 only, every seventh day (55 days), 8 members for both models.}
\label{tab:ci}
\footnotesize
\begin{tabularx}{\textwidth}{@{}llcYY@{}}
\toprule
\textbf{Baseline B} & \textbf{Var.} & \textbf{A $-$ B} [95\% CI]\arrdown & \textbf{Relative}\arrdown & \textbf{Days A better}\arrup \\
\midrule
\multirow{3}{*}{Regression, same inputs} & TG & $-0.396$ [$-0.416$, $-0.369$] & $-28.2$\% & 96\% \\
 & TX & $-0.232$ [$-0.239$, $-0.224$] & $-30.6$\% & 100\% \\
 & TN & $-0.296$ [$-0.311$, $-0.280$] & $-29.9$\% & 100\% \\
\addlinespace[3pt]
\multirow{3}{*}{FNO flow (124M)} & TG & $-0.169$ [$-0.190$, $-0.149$] & $-14.4$\% & 79\% \\
 & TX & $-0.109$ [$-0.115$, $-0.103$] & $-17.2$\% & 97\% \\
 & TN & $-0.143$ [$-0.149$, $-0.135$] & $-17.1$\% & 99\% \\
\addlinespace[3pt]
\multirow{3}{*}{BCSD-style} & TG & $-0.923$ [$-0.969$, $-0.858$] & $-47.8$\% & 98\% \\
 & TX & $-0.472$ [$-0.496$, $-0.449$] & $-47.3$\% & 100\% \\
 & TN & $-0.596$ [$-0.622$, $-0.569$] & $-46.2$\% & 100\% \\
\addlinespace[3pt]
\varrow & TG & $-0.026$ [$-0.028$, $-0.024$] & $-2.5$\% & 95\% \\
\varrow & TX & $-0.008$ [$-0.009$, $-0.008$] & $-1.5$\% & 91\% \\
\varrow \multirow{-3}{*}{\method, 10 Euler steps} & TN & $-0.009$ [$-0.010$, $-0.008$] & $-1.3$\% & 88\% \\
\addlinespace[3pt]
\multirow{3}{*}{CorrDiff-style$^\dagger$} & TG & $-0.224$ [$-0.282$, $-0.070$] & $-17.4$\% & 67\% \\
 & TX & $-0.027$ [$-0.037$, $-0.012$] & $-4.8$\% & 65\% \\
 & TN & $-0.035$ [$-0.045$, $-0.013$] & $-4.8$\% & 76\% \\
\botrule
\end{tabularx}
\end{table}

\begin{table}[!htbp]
\caption{\textbf{Noise prior ablation.} Same setup, only the noise differs; 2019, 16 members, 10 Euler steps. Log score: exact negative log-likelihood per pixel, joint over TG, TX and TN (every seventh day). Bold: best value; green: final configuration; blue: other priors.}
\label{tab:ed_noise}
\footnotesize
\begin{tabularx}{\textwidth}{@{}lYYYcc@{}}
\toprule
\multirow{2}{*}{\textbf{Noise prior}} & \multicolumn{3}{c}{\textbf{CRPS} (\degC)\arrdown} & \textbf{Spread/skill}\toone & \textbf{Log score}\arrdown \\
\cmidrule(lr){2-4}\cmidrule(lr){5-5}\cmidrule(l){6-6}
 & TG & TX & TN & TG & (nats) \\
\midrule
\varrow i.i.d. & 1.037 & 0.536 & 0.700 & 0.50 & 4.65 \\
\varrow RBF, 0.5$^\circ$, 64 features & 1.074 & 0.538 & 0.705 & 0.45 & 4.90 \\
\varrow Power law, 2--500\,km & 1.022 & 0.534 & \textbf{0.698} & 0.51 & 4.67 \\
\varrow Power law, 20--500\,km & \textbf{0.992} & 0.534 & 0.702 & \textbf{0.61} & \textbf{4.57} \\
\oursrow \textbf{Power law, 8--500\,km (final)} & 1.034 & \textbf{0.530} & 0.699 & 0.51 & 4.71 \\
\botrule
\end{tabularx}
\end{table}

\begin{table}[!htbp]
\caption{\textbf{Extremes.} Brier scores for hot (above the local 95th percentile) and cold (below the 5th) events, 2019, 10 Euler steps for \method. Bold: best value; green: \method.}
\label{tab:ed_extremes}
\footnotesize
\begin{tabularx}{\textwidth}{@{}lYYYYYY@{}}
\toprule
\multirow{2}{*}{\textbf{Model}} & \multicolumn{3}{c}{\textbf{Hot}\arrdown} & \multicolumn{3}{c}{\textbf{Cold}\arrdown} \\
\cmidrule(lr){2-4}\cmidrule(l){5-7}
 & TG & TX & TN & TG & TX & TN \\
\midrule
BCSD-style & 0.0588 & 0.0317 & 0.0322 & 0.0428 & 0.0291 & 0.0358 \\
Regression & 0.0432 & 0.0246 & 0.0234 & 0.0280 & 0.0159 & 0.0248 \\
FNO flow (124M) & 0.0382 & 0.0212 & 0.0205 & 0.0243 & 0.0146 & 0.0222 \\
\oursrow \textbf{\method} & \textbf{0.0330} & \textbf{0.0185} & \textbf{0.0171} & \textbf{0.0214} & \textbf{0.0119} & \textbf{0.0188} \\
\botrule
\end{tabularx}
\end{table}

\begin{table}[!htbp]
\caption{\textbf{Inference-time studies.} Every seventh day of 2019, 8 members, \method{} with Euler sampling unless noted. Cell-mean columns: mean absolute difference (\degC) between daily 28\,km cell means of the generated TG and those of NGCD or ERA5; the ERA5 column is a diagnostic rather than a score. Bold: best value; green: sampler used for the main results; blue: other settings.}
\label{tab:ed_infer}
\footnotesize
\begin{tabularx}{\textwidth}{@{}lYYYYY@{}}
\toprule
\multirow{2}{*}{\textbf{Setting}} & \multicolumn{3}{c}{\textbf{CRPS} (\degC)\arrdown} & \multicolumn{2}{c}{\textbf{TG cell means} (\degC)} \\
\cmidrule(lr){2-4}\cmidrule(l){5-6}
 & TG & TX & TN & vs NGCD\arrdown & vs ERA5 \\
\midrule
\grouprow{6}{Consistency with the coarse field (10 Euler steps)}
\varrow Unconstrained (reference) & 1.092 & 0.531 & 0.704 & \textbf{1.61} & 1.88 \\
\varrow Exact ERA5 cell-mean constraint & 1.743 & 0.533 & 0.706 & 1.98 & 0.00 \\
\varrow Exact, monthly debiased & 1.682 & 0.533 & 0.711 & 1.92 & 0.59 \\
\midrule
\grouprow{6}{Sampler (unconstrained)}
\varrow 2 Euler steps & 1.272 & 0.632 & 0.836 & & \\
\varrow 5 Euler steps & 1.130 & 0.552 & 0.729 & & \\
\varrow 20 Euler steps & 1.077 & 0.524 & 0.697 & & \\
\varrow Flow map, 1 step & 1.067 & 0.524 & 0.696 & & \\
\oursrow \textbf{Flow map, 2 steps} & \textbf{1.064} & \textbf{0.523} & \textbf{0.694} & & \\
\botrule
\end{tabularx}
\end{table}

\begin{table}[!htbp]
\caption{\textbf{Generalization to an unseen region.} CRPS (\degC) in 2019 inside the held-out region (northern Fennoscandia, latitude $\geq 66.5^\circ$N and longitude $\geq 24^\circ$E) and in the rest of the domain. Bold: better of the two models; green: model trained on all pixels; blue: model that never sampled the region.}
\label{tab:ed_holdout}
\footnotesize
\begin{tabularx}{\textwidth}{@{}lYYYYYY@{}}
\toprule
\multirow{2}{*}{\textbf{Model}} & \multicolumn{3}{c}{\textbf{Held-out region}\arrdown} & \multicolumn{3}{c}{\textbf{Rest of domain}\arrdown} \\
\cmidrule(lr){2-4}\cmidrule(l){5-7}
 & TG & TX & TN & TG & TX & TN \\
\midrule
\varrow Region never sampled in training & \textbf{1.272} & 0.606 & 0.965 & \textbf{0.969} & 0.528 & 0.682 \\
\oursrow \textbf{Trained on all pixels} & 1.305 & \textbf{0.580} & \textbf{0.901} & 1.004 & \textbf{0.525} & \textbf{0.677} \\
\botrule
\end{tabularx}
\end{table}

\begin{table}[!htbp]
\caption{\textbf{Component ablation.} One change at a time against the matching reference; 2019, every day, fair CRPS (\degC), 10 Euler steps, 16 members unless noted. The static-descriptor and noise rows use a reference with power-law noise over 2--500\,km, the prior with which the model without descriptors was trained. Bold: best value within each block; green: final configuration; blue: other variants. $^*$8 members.}
\label{tab:ed_ablation}
\footnotesize
\begin{tabularx}{\textwidth}{@{}lYYYYYY@{}}
\toprule
\multirow{2}{*}{\textbf{Setting}} & \multicolumn{3}{c}{\textbf{CRPS} (\degC)\arrdown} & \multicolumn{3}{c}{\textbf{Change vs reference} (\%)} \\
\cmidrule(lr){2-4}\cmidrule(l){5-7}
 & TG & TX & TN & TG & TX & TN \\
\midrule
\grouprow{7}{Reference: final configuration}
\oursrow \textbf{\method{} (final configuration)} & \textbf{1.034} & \textbf{0.530} & \textbf{0.699} & \multicolumn{3}{c}{reference} \\
\varrow \quad FNO encoder (124M) instead of Flower & 1.195 & 0.639 & 0.837 & $+15.6$ & $+20.6$ & $+19.7$ \\
\midrule
\grouprow{7}{Reference: power law over 2--500\,km, with static descriptors}
\varrow Power law 2--500\,km, static descriptors & 1.022 & \textbf{0.534} & \textbf{0.698} & \multicolumn{3}{c}{reference} \\
\varrow \quad without static descriptors$^*$ & \textbf{0.987} & 0.548 & 0.731 & $-3.4$ & $+2.6$ & $+4.7$ \\
\varrow \quad RBF noise instead of power law & 1.074 & 0.538 & 0.705 & $+5.1$ & $+0.7$ & $+1.0$ \\
\botrule
\end{tabularx}
\end{table}

\begin{table}[!htbp]
\caption{\textbf{The same configuration transfers to precipitation.} Test year 2019, all pixels; \method{} with 16 members; the regression model uses the same network. CRPS, RMSE and bias in mm\,day$^{-1}$. Frequency bias (1 = correct number of events) and Brier score of the event probability for daily totals above 0.1\,mm (wet days), 10\,mm and 20\,mm. Bold: best value; green: \method; blue: its zero-noise estimate.}
\label{tab:precip}
\footnotesize
\setlength{\tabcolsep}{3pt}%
\begin{tabularx}{\textwidth}{@{}lcccYYYYYY@{}}
\toprule
 & & & & \multicolumn{3}{c}{\textbf{Freq.\ bias}\toone} & \multicolumn{3}{c}{\textbf{Brier score}\arrdown} \\
\cmidrule(lr){5-7}\cmidrule(l){8-10}
\textbf{Model} & \textbf{CRPS}\arrdown & \textbf{RMSE}\arrdown & \textbf{Bias}\,($\to$0) & $>$0.1 & $>$10 & $>$20 & $>$0.1 & $>$10 & $>$20 \\
\midrule
ERA5, bilinear & 2.207 & 4.26 & $+$0.42 & 1.27 & 1.10 & \textbf{0.83} & 0.270 & 0.074 & 0.018 \\
Regression & 1.651 & 3.52 & $-$0.42 & 1.50 & 0.52 & 0.29 & 0.309 & 0.053 & 0.013 \\
\varrow \method, zero-noise ODE & 1.589 & 3.52 & $-$0.57 & 1.13 & 0.50 & 0.24 & 0.219 & 0.052 & 0.013 \\
\oursrow \textbf{\method} & \textbf{1.174} & \textbf{3.51} & \textbf{$-$0.07} & \textbf{1.09} & \textbf{0.94} & 0.79 & \textbf{0.161} & \textbf{0.042} & \textbf{0.011} \\
\botrule
\end{tabularx}
\end{table}

\clearpage
\section*{Supplementary Information}\label{sec:si}
\renewcommand{\thesubsection}{S\arabic{subsection}}%
\setcounter{subsection}{0}%

\subsection{Related work}\label{si:related}

\textbf{Statistical and dynamical downscaling.} Dynamical downscaling nests a regional climate model in a global model or reanalysis and resolves the fine-scale physics explicitly, at a cost that limits ensemble sizes. Statistical downscaling learns a transfer function from coarse to fine fields\cite{maraun2018statistical}. BCSD\cite{wood2004bcsd} interpolates a bias-corrected coarse anomaly and rescales it with a fine-scale climatology; it is routinely used to downscale climate projections and serves here as the statistical baseline. Deep regression networks\cite{vandal2017deepsd,rampal2024review} learn richer transfer functions but predict a conditional mean and therefore under-represent fine-scale variability and extremes.

\textbf{Generative downscaling.} Generative adversarial networks were the first to produce realistic, stochastic kilometre-scale fields\cite{leinonen2021gan,harris2022gan}. Diffusion models\cite{ho2020ddpm,song2021sde,karras2022edm} have since become the dominant approach\cite{watt2024diffusion,mardani2025corrdiff}. CorrDiff\cite{mardani2025corrdiff} predicts the mean with a UNet and models the residual with a diffusion model, recovering the spectra of the target. GenFocal\cite{wan2024genfocal} combines a rectified-flow debiasing step with a conditional diffusion super-resolution step to downscale climate projections without temporally aligned training pairs. These models operate on a fixed grid, sample with tens of network evaluations and do not provide a tractable likelihood at the scale of a full field. \method{} differs in all three respects and is complementary to approaches such as GenFocal: its pointwise, resolution-free design could replace the super-resolution stage of such pipelines.

\textbf{Flow matching, functional generative models and flow maps.} Flow matching and stochastic interpolants\cite{lipman2023flow,liu2023rectified,albergo2023interpolants} learn the velocity of a transport between noise and data by regression. Functional flow matching\cite{kerrigan2024ffm} defines this transport between Gaussian processes and data supported on functions, so that samples can be evaluated on arbitrary discretizations. Continuous normalizing flows\cite{chen2018node,grathwohl2019ffjord} provide exact likelihoods through the instantaneous change of variables but require stochastic trace estimators\cite{hutchinson1989trace} in high dimension. Flow maps and shortcut models\cite{boffi2024flowmap,frans2025shortcut,ouyang2026scallop} learn the integral of the velocity between two times to sample in few steps. Neural operators\cite{li2021fno,flower} learn maps between function spaces and serve here as the encoder (Flower) and as a baseline (FNO).

\subsection{Functional consistency of the pointwise flow}\label{si:consistency}

Let $z$ be a Gaussian random field with continuous covariance and let $\phi_{c(s)}$ denote the time-1 flow map of $v_\theta(\cdot, c(s), \cdot)$. For any finite point set $\{s_1,\dots,s_n\}$, \method{} generates the vector $(\phi_{c(s_1)}(z(s_1)),\dots,\phi_{c(s_n)}(z(s_n)))$. Its law $\mu_{s_1\dots s_n}$ is the push-forward of the finite-dimensional distribution of $z$ under the product map $\phi_{c(s_1)}\times\dots\times\phi_{c(s_n)}$. Permuting the points permutes the coordinates of both $z$ and the product map, and marginalizing a point removes one coordinate of $z$ together with its map; the family $\{\mu_{s_1\dots s_n}\}$ therefore inherits the permutation and marginalization consistency of the finite-dimensional distributions of $z$. By Kolmogorov's extension theorem it defines a stochastic process $x(s)$ on the domain, which is what allows the same model to be queried on any grid or set of points. Because the maps act on single coordinates, the copula of $(x(s_1),\dots,x(s_n))$ given the condition equals that of $(z(s_1),\dots,z(s_n))$ up to the monotone per-coordinate transformations: the network cannot introduce spatial dependence that is absent from $z$. Finally, the flow-matching loss is an expectation over locations of per-location terms, so the loss evaluated on points drawn at random from the domain is an unbiased estimate of the loss on the whole field.

\subsection{Implementation details of the CorrDiff-style baseline}\label{si:baselines}

The diffusion stage uses Karras preconditioning ($\sigma_\mathrm{data}=0.5$), log-normal noise levels and a deterministic Heun sampler with 18 steps. Both UNets have channel multipliers 1-2-4-4 (base 64), two residual blocks per level with FiLM noise embedding and no attention, and are therefore fully convolutional. Their inputs are bilinearly interpolated ERA5 fields, the 17 static descriptors, day of year, normalized latitude and longitude and a land mask, all at 1\,km. Both stages are trained on random 128$\times$128\,km patches, for 60{,}000 steps (regression, 3.1\,h) and 150{,}000 steps (diffusion, 10.0\,h), with checkpoints selected on a fixed validation-patch loss. As in the original CorrDiff, inference is tile-wise: both UNets are applied to overlapping 128$\times$128\,km tiles (stride 112\,km, linear blending over the overlap) at every regression and denoising call, so that the group-normalization statistics match those seen in training. Applying the UNets to the whole domain at once, which includes about 40\% sea, changes these statistics; in a preliminary test this degraded the regression stage from a TG mean absolute error of 1.62 to 2.11\degC, so we report only the tile-wise results.

\subsection{Further analyses}\label{si:further}

\textbf{Fine-scale detail within reanalysis cells.} The departures of \method{} samples from the mean of their 0.25$^\circ$ cell are too wide: pooled over 20{,}000 fixed pixels and every day of 2019, their interquartile range is 2.4 times that of NGCD, whereas the other methods lie within 0.95--1.11 times it. This is the excess power at 10--50\,km visible in the spectra (Fig.~\ref{fig:texture}): a stationary noise prior adds sub-cell variability everywhere, including where the observed field is smooth. A prior whose band weights depend on terrain roughness is a natural remedy that keeps the noise a Gaussian random field.

\textbf{Unphysical cold samples.} Pooled over 20{,}000 fixed pixels and every day of 2019, the coldest \method{} sample of TN is $-59.5$\degC, against $-39.7$\degC{} in NGCD at the same pixels; 413 of 7.3 million pixel-days are below $-40$\degC{} and 13 below $-50$\degC, typically on days when NGCD shows moderate frost at those pixels (FNO flow: 12 values below $-40$\degC; BCSD-style: 27; regression: none). These outliers barely affect the scores but matter for applications that use individual samples; the unbounded standardized anomaly in the inverse transform is the natural place to intervene.

\textbf{Static descriptors.} Without static descriptors, the standardized output of the model is nearly flat at the pixel scale, even where NGCD shows weather-dependent structure such as cold-air pooling: in western Norway, the pixel-scale roughness of standardized anomalies is then 0.001, against 0.04--0.06 in NGCD. Adding the descriptors lowers the 2019 CRPS for TX and TN by 3 and 5\% (0.548 to 0.534 and 0.731 to 0.698\degC) and raises that of TG slightly (0.99 to 1.02\degC), consistent with TG being the variable least determined by the local climate.

\textbf{Daily mean temperature in NGCD.} In NGCD 2019, TG differs from (TX$+$TN)/2 by 1.70\degC{} on average (spatial standard deviation 1.8\degC; 2.2\degC{} in winter), with day-to-day shifts of the domain mean of up to 3.3\degC, because TG is defined and interpolated independently of TX and TN. This variability is not determined by TX, TN or ERA5 and explains why TG errors exceed those of TX and TN for every model.

\textbf{Held-out region.} For TG, the model that never sampled northern Fennoscandia is 0.03\degC{} better both inside and outside the region, so withholding the region has no detectable effect. For TX and TN, the region is slightly worse (by 4 and 7\%), more than the differences outside it; confirming this small penalty would require repeated training runs. The region is intrinsically harder than the rest of the domain (TG CRPS 1.31 against 1.00 even when it is seen), so most of its error reflects its climate rather than its absence from training.

\end{appendices}

\clearpage
\bibliography{main}

\begin{thebibliography}{10}
\expandafter\ifx\csname url\endcsname\relax
  \def\url#1{\burl{#1}}\fi
\expandafter\ifx\csname urlprefix\endcsname\relax\def\urlprefix{URL }\fi
\providecommand{\bibinfo}[2]{#2}
\providecommand{\eprint}[2][]{\url{#2}}
\providecommand{\doi}[1]{\url{https://doi.org/#1}}
\bibcommenthead

\bibitem{fischer2021record}
\bibinfo{author}{Fischer, E.~M.}, \bibinfo{author}{Sippel, S.} \& \bibinfo{author}{Knutti, R.}
\newblock \bibinfo{title}{Increasing probability of record-shattering climate extremes}.
\newblock \emph{\bibinfo{journal}{Nature Climate Change}} \textbf{\bibinfo{volume}{11}}, \bibinfo{pages}{689--695} (\bibinfo{year}{2021}).

\bibitem{hersbach2020era5}
\bibinfo{author}{Hersbach, H.}, \bibinfo{author}{Bell, B.}, \bibinfo{author}{Berrisford, P.} \emph{et~al.}
\newblock \bibinfo{title}{The {ERA5} global reanalysis}.
\newblock \emph{\bibinfo{journal}{Quarterly Journal of the Royal Meteorological Society}} \textbf{\bibinfo{volume}{146}}, \bibinfo{pages}{1999--2049} (\bibinfo{year}{2020}).

\bibitem{maraun2018statistical}
\bibinfo{author}{Maraun, D.} \& \bibinfo{author}{Widmann, M.}
\newblock \emph{\bibinfo{title}{Statistical Downscaling and Bias Correction for Climate Research}}  (\bibinfo{publisher}{Cambridge University Press}, \bibinfo{year}{2018}).

\bibitem{wan2024genfocal}
\bibinfo{author}{Wan, Z.~Y.} \emph{et~al.}
\newblock \bibinfo{title}{Regional climate risk assessment from climate models using probabilistic machine learning}.
\newblock \emph{\bibinfo{journal}{arXiv preprint arXiv:2412.08079}}  (\bibinfo{year}{2024}).

\bibitem{wood2004bcsd}
\bibinfo{author}{Wood, A.~W.}, \bibinfo{author}{Leung, L.~R.}, \bibinfo{author}{Sridhar, V.} \& \bibinfo{author}{Lettenmaier, D.~P.}
\newblock \bibinfo{title}{Hydrologic implications of dynamical and statistical approaches to downscaling climate model outputs}.
\newblock \emph{\bibinfo{journal}{Climatic Change}} \textbf{\bibinfo{volume}{62}}, \bibinfo{pages}{189--216} (\bibinfo{year}{2004}).

\bibitem{vandal2017deepsd}
\bibinfo{author}{Vandal, T.} \emph{et~al.}
\newblock \emph{\bibinfo{title}{{DeepSD}: generating high resolution climate change projections through single image super-resolution}}, \bibinfo{pages}{1663--1672} (\bibinfo{year}{2017}).

\bibitem{rampal2024review}
\bibinfo{author}{Rampal, N.}, \bibinfo{author}{Hobeichi, S.}, \bibinfo{author}{Gibson, P.~B.} \emph{et~al.}
\newblock \bibinfo{title}{Enhancing regional climate downscaling through advances in machine learning}.
\newblock \emph{\bibinfo{journal}{Artificial Intelligence for the Earth Systems}} \textbf{\bibinfo{volume}{3}} (\bibinfo{year}{2024}).

\bibitem{leinonen2021gan}
\bibinfo{author}{Leinonen, J.}, \bibinfo{author}{Nerini, D.} \& \bibinfo{author}{Berne, A.}
\newblock \bibinfo{title}{Stochastic super-resolution for downscaling time-evolving atmospheric fields with a generative adversarial network}.
\newblock \emph{\bibinfo{journal}{IEEE Transactions on Geoscience and Remote Sensing}} \textbf{\bibinfo{volume}{59}}, \bibinfo{pages}{7211--7223} (\bibinfo{year}{2021}).

\bibitem{harris2022gan}
\bibinfo{author}{Harris, L.}, \bibinfo{author}{McRae, A. T.~T.}, \bibinfo{author}{Chantry, M.}, \bibinfo{author}{Dueben, P.~D.} \& \bibinfo{author}{Palmer, T.~N.}
\newblock \bibinfo{title}{A generative deep learning approach to stochastic downscaling of precipitation forecasts}.
\newblock \emph{\bibinfo{journal}{Journal of Advances in Modeling Earth Systems}} \textbf{\bibinfo{volume}{14}}, \bibinfo{pages}{e2022MS003120} (\bibinfo{year}{2022}).

\bibitem{watt2024diffusion}
\bibinfo{author}{Watt, R.~A.} \& \bibinfo{author}{Mansfield, L.~A.}
\newblock \bibinfo{title}{Generative diffusion-based downscaling for climate} (\bibinfo{year}{2024}).
\newblock \bibinfo{note}{ArXiv:2404.17752}.

\bibitem{mardani2025corrdiff}
\bibinfo{author}{Mardani, M.} \emph{et~al.}
\newblock \bibinfo{title}{Residual corrective diffusion modeling for km-scale atmospheric downscaling}.
\newblock \emph{\bibinfo{journal}{Communications Earth \& Environment}} \textbf{\bibinfo{volume}{6}}, \bibinfo{pages}{124} (\bibinfo{year}{2025}).

\bibitem{ravuri2021dgmr}
\bibinfo{author}{Ravuri, S.}, \bibinfo{author}{Lenc, K.}, \bibinfo{author}{Willson, M.} \emph{et~al.}
\newblock \bibinfo{title}{Skilful precipitation nowcasting using deep generative models of radar}.
\newblock \emph{\bibinfo{journal}{Nature}} \textbf{\bibinfo{volume}{597}}, \bibinfo{pages}{672--677} (\bibinfo{year}{2021}).

\bibitem{price2025gencast}
\bibinfo{author}{Price, I.}, \bibinfo{author}{Sanchez-Gonzalez, A.}, \bibinfo{author}{Alet, F.} \emph{et~al.}
\newblock \bibinfo{title}{Probabilistic weather forecasting with machine learning}.
\newblock \emph{\bibinfo{journal}{Nature}} \textbf{\bibinfo{volume}{637}}, \bibinfo{pages}{84--90} (\bibinfo{year}{2025}).

\bibitem{karras2022edm}
\bibinfo{author}{Karras, T.}, \bibinfo{author}{Aittala, M.}, \bibinfo{author}{Aila, T.} \& \bibinfo{author}{Laine, S.}
\newblock \emph{\bibinfo{title}{Elucidating the design space of diffusion-based generative models}}, Vol.~\bibinfo{volume}{35} (\bibinfo{year}{2022}).

\bibitem{hutchinson1989trace}
\bibinfo{author}{Hutchinson, M.~F.}
\newblock \bibinfo{title}{A stochastic estimator of the trace of the influence matrix for {Laplacian} smoothing splines}.
\newblock \emph{\bibinfo{journal}{Communications in Statistics -- Simulation and Computation}} \textbf{\bibinfo{volume}{18}}, \bibinfo{pages}{1059--1076} (\bibinfo{year}{1989}).

\bibitem{grathwohl2019ffjord}
\bibinfo{author}{Grathwohl, W.}, \bibinfo{author}{Chen, R. T.~Q.}, \bibinfo{author}{Bettencourt, J.}, \bibinfo{author}{Sutskever, I.} \& \bibinfo{author}{Duvenaud, D.}
\newblock \emph{\bibinfo{title}{{FFJORD}: free-form continuous dynamics for scalable reversible generative models}} (\bibinfo{year}{2019}).

\bibitem{lipman2023flow}
\bibinfo{author}{Lipman, Y.}, \bibinfo{author}{Chen, R. T.~Q.}, \bibinfo{author}{Ben-Hamu, H.}, \bibinfo{author}{Nickel, M.} \& \bibinfo{author}{Le, M.}
\newblock \emph{\bibinfo{title}{Flow matching for generative modeling}} (\bibinfo{year}{2023}).

\bibitem{liu2023rectified}
\bibinfo{author}{Liu, X.}, \bibinfo{author}{Gong, C.} \& \bibinfo{author}{Liu, Q.}
\newblock \emph{\bibinfo{title}{Flow straight and fast: learning to generate and transfer data with rectified flow}} (\bibinfo{year}{2023}).

\bibitem{albergo2023interpolants}
\bibinfo{author}{Albergo, M.~S.} \& \bibinfo{author}{Vanden-Eijnden, E.}
\newblock \emph{\bibinfo{title}{Building normalizing flows with stochastic interpolants}} (\bibinfo{year}{2023}).

\bibitem{kerrigan2024ffm}
\bibinfo{author}{Kerrigan, G.}, \bibinfo{author}{Migliorini, G.} \& \bibinfo{author}{Smyth, P.}
\newblock \emph{\bibinfo{title}{Functional flow matching}}, Vol. \bibinfo{volume}{238} of \emph{\bibinfo{series}{PMLR}} (\bibinfo{year}{2024}).

\bibitem{boffi2024flowmap}
\bibinfo{author}{Boffi, N.~M.}, \bibinfo{author}{Albergo, M.~S.} \& \bibinfo{author}{Vanden-Eijnden, E.}
\newblock \bibinfo{title}{Flow map matching} (\bibinfo{year}{2024}).
\newblock \bibinfo{note}{ArXiv:2406.07507}.

\bibitem{ngcd}
\bibinfo{author}{{Copernicus Climate Change Service}}.
\newblock \bibinfo{title}{Nordic gridded temperature and precipitation data from 1961 to present derived from in-situ observations}.
\newblock \bibinfo{howpublished}{Copernicus Climate Change Service (C3S) Climate Data Store} (\bibinfo{year}{2021}).

\bibitem{li2021fno}
\bibinfo{author}{Li, Z.} \emph{et~al.}
\newblock \emph{\bibinfo{title}{Fourier neural operator for parametric partial differential equations}} (\bibinfo{year}{2021}).

\bibitem{flower}
\bibinfo{author}{Muser, T.}, \bibinfo{author}{Spitzer, A.}, \bibinfo{author}{Lassas, M.}, \bibinfo{author}{De~Hoop, M.~V.} \& \bibinfo{author}{Dokmani\'{c}, I.}
\newblock \emph{\bibinfo{title}{Flowers: A warp drive for neural {PDE} solvers}}, Vol. \bibinfo{volume}{306} of \emph{\bibinfo{series}{Proceedings of Machine Learning Research}}, \bibinfo{pages}{91390--91418} (\bibinfo{publisher}{PMLR}, \bibinfo{year}{2026}).
\newblock \urlprefix\url{https://proceedings.mlr.press/v306/muser26a.html}.

\bibitem{chen2021liif}
\bibinfo{author}{Chen, Y.}, \bibinfo{author}{Liu, S.} \& \bibinfo{author}{Wang, X.}
\newblock \emph{\bibinfo{title}{Learning continuous image representation with local implicit image function}}, \bibinfo{pages}{8628--8638} (\bibinfo{year}{2021}).

\bibitem{chen2018node}
\bibinfo{author}{Chen, R. T.~Q.}, \bibinfo{author}{Rubanova, Y.}, \bibinfo{author}{Bettencourt, J.} \& \bibinfo{author}{Duvenaud, D.}
\newblock \emph{\bibinfo{title}{Neural ordinary differential equations}}, Vol.~\bibinfo{volume}{31} (\bibinfo{year}{2018}).

\bibitem{frans2025shortcut}
\bibinfo{author}{Frans, K.}, \bibinfo{author}{Hafner, D.}, \bibinfo{author}{Levine, S.} \& \bibinfo{author}{Abbeel, P.}
\newblock \emph{\bibinfo{title}{One step diffusion via shortcut models}} (\bibinfo{year}{2025}).

\bibitem{kunsch1989bootstrap}
\bibinfo{author}{K{\"u}nsch, H.~R.}
\newblock \bibinfo{title}{The jackknife and the bootstrap for general stationary observations}.
\newblock \emph{\bibinfo{journal}{The Annals of Statistics}} \textbf{\bibinfo{volume}{17}}, \bibinfo{pages}{1217--1241} (\bibinfo{year}{1989}).

\bibitem{lussana2019senorge}
\bibinfo{author}{Lussana, C.}, \bibinfo{author}{Tveito, O.~E.}, \bibinfo{author}{Dobler, A.} \& \bibinfo{author}{Tunheim, K.}
\newblock \bibinfo{title}{senorge\_2018, daily precipitation, and temperature datasets over {Norway}}.
\newblock \emph{\bibinfo{journal}{Earth System Science Data}} \textbf{\bibinfo{volume}{11}}, \bibinfo{pages}{1531--1551} (\bibinfo{year}{2019}).

\bibitem{rahimi2007rff}
\bibinfo{author}{Rahimi, A.} \& \bibinfo{author}{Recht, B.}
\newblock \emph{\bibinfo{title}{Random features for large-scale kernel machines}}, Vol.~\bibinfo{volume}{20} (\bibinfo{year}{2007}).

\bibitem{loshchilov2019adamw}
\bibinfo{author}{Loshchilov, I.} \& \bibinfo{author}{Hutter, F.}
\newblock \emph{\bibinfo{title}{Decoupled weight decay regularization}} (\bibinfo{year}{2019}).

\bibitem{ouyang2026scallop}
\bibinfo{author}{OuYang, R.} \emph{et~al.}
\newblock \bibinfo{title}{Few-step {Boltzmann} generators via scalable likelihood flow maps} (\bibinfo{year}{2026}).
\newblock \bibinfo{note}{ArXiv:2606.29110}.

\bibitem{ferro2014fair}
\bibinfo{author}{Ferro, C. A.~T.}
\newblock \bibinfo{title}{Fair scores for ensemble forecasts}.
\newblock \emph{\bibinfo{journal}{Quarterly Journal of the Royal Meteorological Society}} \textbf{\bibinfo{volume}{140}}, \bibinfo{pages}{1917--1923} (\bibinfo{year}{2014}).

\bibitem{gneiting2007scoring}
\bibinfo{author}{Gneiting, T.} \& \bibinfo{author}{Raftery, A.~E.}
\newblock \bibinfo{title}{Strictly proper scoring rules, prediction, and estimation}.
\newblock \emph{\bibinfo{journal}{Journal of the American Statistical Association}} \textbf{\bibinfo{volume}{102}}, \bibinfo{pages}{359--378} (\bibinfo{year}{2007}).

\bibitem{hamill2001rank}
\bibinfo{author}{Hamill, T.~M.}
\newblock \bibinfo{title}{Interpretation of rank histograms for verifying ensemble forecasts}.
\newblock \emph{\bibinfo{journal}{Monthly Weather Review}} \textbf{\bibinfo{volume}{129}}, \bibinfo{pages}{550--560} (\bibinfo{year}{2001}).

\bibitem{ebert2001pm}
\bibinfo{author}{Ebert, E.~E.}
\newblock \bibinfo{title}{Ability of a poor man's ensemble to predict the probability and distribution of precipitation}.
\newblock \emph{\bibinfo{journal}{Monthly Weather Review}} \textbf{\bibinfo{volume}{129}}, \bibinfo{pages}{2461--2480} (\bibinfo{year}{2001}).

\bibitem{ho2020ddpm}
\bibinfo{author}{Ho, J.}, \bibinfo{author}{Jain, A.} \& \bibinfo{author}{Abbeel, P.}
\newblock \emph{\bibinfo{title}{Denoising diffusion probabilistic models}}, Vol.~\bibinfo{volume}{33} (\bibinfo{year}{2020}).

\bibitem{song2021sde}
\bibinfo{author}{Song, Y.} \emph{et~al.}
\newblock \emph{\bibinfo{title}{Score-based generative modeling through stochastic differential equations}} (\bibinfo{year}{2021}).

\end{thebibliography}

\end{document}